\documentclass[default, natbib, authoryear]{sn-jnl}
\usepackage{graphicx}%
\usepackage{multirow}%
\usepackage{amsmath,amssymb,amsfonts}%
\usepackage{amsthm}%
\usepackage{mathtools}%
\usepackage{mathrsfs}%
\usepackage{bbm}
\usepackage[title]{appendix}%
\usepackage{xcolor}%
\usepackage{textcomp}%
\usepackage{manyfoot}%
\usepackage{booktabs}%
\usepackage{algorithm}%
\usepackage{algorithmicx}%
\usepackage{algpseudocode}%
\usepackage{listings}%
\usepackage{microtype}%
\usepackage{placeins}
\usepackage{subfig}%
\usepackage{tikz}%
\usetikzlibrary{positioning, calc, decorations.pathreplacing}%

\usepackage{anyfontsize}

\setcitestyle{round}

\usepackage{color}
\definecolor{darkred}{rgb}{0.6,0.0,0.0}
\definecolor{darkgreen}{rgb}{0,0.50,0}
\definecolor{lightblue}{rgb}{0.0,0.42,0.91}
\definecolor{orange}{rgb}{0.99,0.48,0.13}
\definecolor{grass}{rgb}{0.18,0.80,0.18}
\definecolor{pink}{rgb}{0.97,0.15,0.45}

\lstdefinestyle{colored}{ %
  basicstyle=\ttfamily,
  backgroundcolor=\color{white},
  commentstyle=\color{green}\itshape,
  keywordstyle=\color{blue}\bfseries\itshape,
  stringstyle=\color{red},
}
\lstdefinelanguage{PythonPlus}[]{Python}{
  morekeywords=[1]{,as,assert,nonlocal,with,yield,self,True,False,None,} % Python builtin
  morekeywords=[2]{,__init__,__add__,__mul__,__div__,__sub__,__call__,__getitem__,__setitem__,__eq__,__ne__,__nonzero__,__rmul__,__radd__,__repr__,__str__,__get__,__truediv__,__pow__,__name__,__future__,__all__,}, % magic methods
  morekeywords=[3]{,object,type,isinstance,copy,deepcopy,zip,enumerate,reversed,list,set,len,dict,tuple,range,xrange,append,execfile,real,imag,reduce,str,repr,}, % common functions
  morekeywords=[4]{,Exception,NameError,IndexError,SyntaxError,TypeError,ValueError,OverflowError,ZeroDivisionError,}, % errors
  morekeywords=[5]{,ode,fsolve,sqrt,exp,sin,cos,arctan,arctan2,arccos,pi, array,norm,solve,dot,arange,isscalar,max,sum,flatten,shape,reshape,find,any,all,abs,plot,linspace,legend,quad,polyval,polyfit,hstack,concatenate,vstack,column_stack,empty,zeros,ones,rand,vander,grid,pcolor,eig,eigs,eigvals,svd,qr,tan,det,logspace,roll,min,mean,cumsum,cumprod,diff,vectorize,lstsq,cla,eye,xlabel,ylabel,squeeze,}, % numpy / math
}

\lstdefinestyle{colorEX}{
  basicstyle=\ttfamily,
  backgroundcolor=\color{white},
  commentstyle=\color{darkgreen}\slshape,
  keywordstyle=\color{blue}\bfseries\itshape,
  keywordstyle=[2]\color{blue}\bfseries,
  keywordstyle=[3]\color{grass},
  keywordstyle=[4]\color{red},
  keywordstyle=[5]\color{orange},
  stringstyle=\color{darkred},
  emphstyle=\color{pink}\underbar,
}

\renewcommand{\vec}{\pmb}
\newcommand{\I}{\mathcal{I}}
\newcommand{\E}{\mathbb{E}}
\DeclareMathOperator*{\argmax}{arg\,max}

\newcommand{\std}[1]{\textsubscript{#1}}
\newcommand{\diff}[2]{\frac{\partial #1}{\partial #2}}

\begin{document}
%%%%%%%%%%%%%%%%%%%%%%%%%%%%%%%%%%%%%%%%%%%%%%%%%%%%%%%%%%%%%%%%%%%%%%%%%
% COVER
%%%%%%%%%%%%%%%%%%%%%%%%%%%%%%%%%%%%%%%%%%%%%%%%%%%%%%%%%%%%%%%%%%%%%%%%%

\title[Sparse GEMINI for joint discriminative clustering and feature selection]{Sparse and Geometry-aware Generalisation of the Mutual Information for Joint Discriminative Clustering and Feature Selection}

\author*[1,2]{\fnm{Louis} \sur{Ohl}}\email{louis.ohl@inria.fr}
\author*[1]{\fnm{Pierre-Alexandre} \sur{Mattei}}\email{pierre-alexandre.mattei@inria.fr}
\author[1]{\fnm{Charles} \sur{Bouveyron}}
\author[2]{\fnm{Micka\"el}\sur{Leclercq}}
\author[2]{\fnm{Arnaud}\sur{Droit}}
\author[1]{\fnm{Fr\'ed\'eric}\sur{Precioso}}

\affil[1]{\orgname{Universit\'e C\^ote d'Azur, Inria, CNRS, Maasai Team}}
\affil[2]{\orgname{CHU de Qu\' ebec Research Centre, Laval University}}

\abstract{Feature selection in clustering is a hard task which involves simultaneously the discovery of relevant clusters as well as relevant variables with respect to these clusters. While feature selection algorithms are often model-based through optimised model selection or strong assumptions on the data distribution, we introduce a discriminative clustering model trying to maximise a geometry-aware generalisation of the mutual information called GEMINI with a simple $\ell_1$ penalty: the Sparse GEMINI. This algorithm avoids the burden of combinatorial feature subset exploration and is easily scalable to high-dimensional data and large amounts of samples while only designing a discriminative clustering model. We demonstrate the performances of Sparse GEMINI on synthetic datasets and large-scale datasets. Our results show that Sparse GEMINI is a competitive algorithm and has the ability to select relevant subsets of variables with respect to the clustering without using relevance criteria or prior hypotheses.}

\maketitle

%%%%%%%%%%%%%%%%%%%%%%%%%%%%%%%%%%%%%%%%%%%%%%%%%%%%%%%%%%%%%%%%%%%%%%%%%
% INTRODUCTION
%%%%%%%%%%%%%%%%%%%%%%%%%%%%%%%%%%%%%%%%%%%%%%%%%%%%%%%%%%%%%%%%%%%%%%%%%

\section{Introduction}
\label{sec:introducion}

It is common that clustering algorithms and supervised models rely on all available features for the best performance. However, as data sets become high-dimensional, clustering algorithms tend to break under the curse of dimensionality~\citep{bouveyron_model-based_2014}, for instance in biological micro-array data where the number of variables outweighs the number of samples~\citep{mclachlan_mixture_2002}. To alleviate this burden, feature selection is a method of choice. Indeed, all features may not always be of interest: some variables can be perceived as relevant or not with respect to the clustering objective. Relevant variables bring information that is useful for the clustering operation, while irrelevant variables do not bring any new knowledge regarding the cluster distribution~\citep{tadesse_bayesian_2005} and redundant variables look relevant but do not bring beneficial knowledge~\citep{maugis_variable_2009}. The challenge of selecting the relevant variables often comes with the burden of combinatorial search in the variable space. Therefore, solutions may be hardly scalable to high-dimensional data~\citep{raftery_variable_2006} or to the number of samples~\citep{witten_framework_2010} when the selection process is part of the model. Therefore reducing the number of variables for learning to a relevant few is of interest, notably in terms of interpretation~\citep{fop_variable_2018}. The necessity of variable selection notably met successful applications in genomics~\citep{marbac_variable_2020}, multi-omics~\citep{meng_mocluster_2016,ramazzotti_multi-omic_2018,shen_integrative_2012}.

Often, integrating the selection process as part of the model will lead to either not scaling well~\citep{solorio-fernandez_review_2020} in terms of number of features~\citep{raftery_variable_2006} or number of samples~\citep{witten_framework_2010} or imposing too constrained decision boundaries due to the nature of strong parametric assumptions. To alleviate both problems, we present the Sparse GEMINI: a model that combines the LassoNet architecture~\citep{lemhadri_lassonet_2021} and the discriminative clustering objective GEMINI~\citep{ohl_generalised_2022, ohl_generalised_2023} for a scalable discriminative clustering with penalised feature selection. The contributions of Sparse GEMINI are:

\begin{itemize}
\item A simple novel scalable algorithm efficiently combining feature selection and discriminative clustering compatible with models ranging from logistic regression to deep neural networks.
\item Demonstrations of performances on multiple synthetic and real datasets including a large-scale transcriptomics dataset.
\item An extension of the work on the generalised mutual information by proposing a package containing the methods of previous work and the Sparse GEMINI model thanks to explicit computations of GEMINI gradients.
\end{itemize}

%%%%%%%%%%%%%%%%%%%%%%%%%%%%%%%%%%%%%%%%%%%%%%%%%%%%%%%%%%%%%%%%%%%%%%%%%
% RELATED WORKS
%%%%%%%%%%%%%%%%%%%%%%%%%%%%%%%%%%%%%%%%%%%%%%%%%%%%%%%%%%%%%%%%%%%%%%%%%

\section{Related works}
\label{sec:related_works}

\def\nodesep{0.1}
\def\basesep{1}
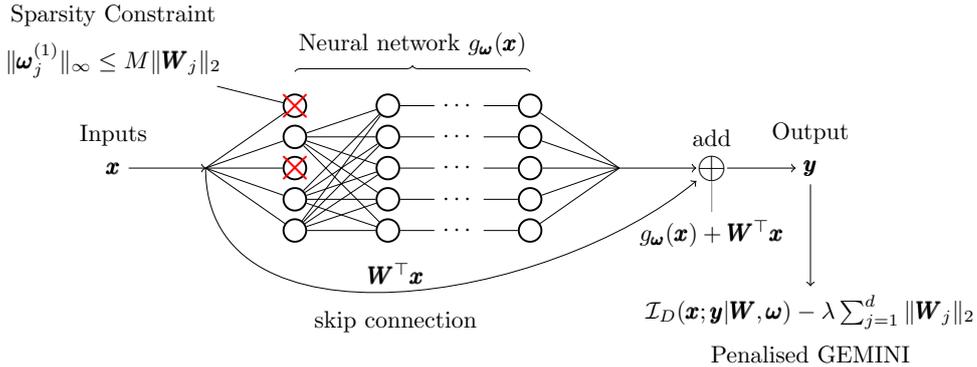
\begin{figure*}[!t]
\centering
\resizebox{\linewidth}{!}{
\begin{tikzpicture}[
    node/.style={circle, draw, thick},
  ]

\node[label={above:Inputs}] (inputs) {$\vec{x}$}; 
  \node[right= of inputs] (l1_entrance) {};
  
  \node[node, right=of l1_entrance] (i3) {};
  \node[red, font=\huge] at (i3) {$\times$};
  \node[node, above=\nodesep of i3] (i2) {};
  \node[node, above=\nodesep of i2] (i1) {};
  \node[red, font=\huge] at (i1) {$\times$};
  \node[node, below=\nodesep of i3] (i4) {};
  \node[node, below=\nodesep of i4] (i5) {};
  
  \foreach \y in {1,...,5}
  \node[node, right=\basesep of i\y] (h\y) {};
  
  \foreach \y in {1,...,5}
  \node[right=0.5*\basesep of h\y] (t\y) {$\cdots$};
  
  \foreach \y in {1,...,5}
  \node[node, right=0.5*\basesep of t\y] (o\y) {};

  \node[right=\basesep of o3] (hidden_entrance) {};
  
  \node[right=\basesep of hidden_entrance, font=\Large, label={above:add}, inner sep=0, pin={-90:$g_{\vec{\omega}}(\vec{x}) + \vec{W}^\top\vec{x}$}] (add) {$\oplus$};
  \node[right=\basesep of add, label={above:Output}] (y_pred) {$\vec{y}$};
  \node[below=1.5*\basesep of y_pred, label={below:Penalised GEMINI}] (gemini) {$\I_D(\vec{x};\vec{y}| \vec{W},\vec{\omega}) - \lambda \sum_{j=1}^d\|\vec{W}_j\|_2$} ;
  
  \foreach \source in {2,4,5}
  \foreach \dest in {1, ..., 5}
  \draw[-] (i\source) -- (h\dest);
    
  \foreach \dest in {1, ..., 5}
  \draw[-] (l1_entrance.center) -- (i\dest);
  
  \foreach \dest in {1, ..., 5}
  \draw[-] (o\dest) -- (hidden_entrance.center);
  
  \foreach \dest in {1, ..., 5}
  \draw[-] (h\dest) -- (t\dest);
  
  \foreach \dest in {1,...,5}
  \draw[-] (t\dest) -- (o\dest);
  
  \draw[->] (hidden_entrance.center) -- (add);

  \draw[<-] (l1_entrance.center) -- (inputs);
  \draw[->] (add) -- (y_pred) node[above, pos=0.8] {};
  \draw[->] (y_pred) -- (gemini);
  
  \draw[decorate, decoration={brace}] ($(i1)+(0,5*\nodesep)$) -- node[midway, above=\nodesep] {Neural network $g_{\vec{\omega}}(\vec{x})$} ($(o1)+(0,5*\nodesep)$);
  
  \node[above=\basesep of inputs, label={above:Sparsity Constraint}] (constraints) {$\|\vec{\omega}^{(1)}_j\|_{\infty} \leq M \|\vec{W}_j\|_2$};
  \draw[-] (constraints) -- (i1);
  
    \draw[->] ($(l1_entrance)-(0,0)$) to[out=-90, in=-140] node[below=1ex, midway, align=center] {skip connection} node[above, midway] {$\vec{W}^\top\vec{x}$} (add);

\end{tikzpicture}}
\caption{Description of the complete Sparse GEMINI model. Through a proximal gradient, clusters learned by GEMINI drop irrelevant features both in a skip connection and an MLP. Setting $M=0$ recovers a sparse unsupervised logistic regression.}\label{fig:sparse_gemini_overview}
\end{figure*}

Feature selection algorithms can be divided into 2 distinct categories~\citep{john_irrelevant_1994, dy_unsupervised_2007}: filter methods and wrapper methods. Filter methods apply in an independent step feature selection using a relevance criterion to eliminate irrelevant features before performing clustering. This can be done, for example, using information theory~\citep{cover_elements_1999} with the SVD-Entropy~\citep{varshavsky_novel_2006} or spectral analysis~\citep{von_luxburg_tutorial_2007, he_laplacian_2005, zhao_spectral_2007}. Those methods are thus easily scalable and quick despite the challenge of defining unsupervised feature relevance~\citep{dy_unsupervised_2007}. Wrapper methods encompass the selection process within the model and exploit their clustering results to guide the feature selection~\citep{solorio-fernandez_review_2020}. Other related works sometimes refer to a third category named hybrid model~\citep{alelyani_feature_2018} or embedded models~\citep{blum_selection_1997} as a compromise between the first two categories.

Although the definition of relevance of a variable is more straightforward for supervised learning, its definition in unsupervised learning clearly impacts the choice of selection criterion for filter methods or distribution design in model-based methods~\citep{fop_variable_2018}. Often, the terms relevant variables, irrelevant variables~\citep{tadesse_bayesian_2005} for the notion of conveying information are used. Others may consider redundant variables as those that bring information already available~\citep{maugis_variable_2009}. Another key difference in models would be to consider whether the informative variables are independent given the cluster membership (local independence) or dependent (global independence from the non-informative variables), although the latter hardly accounts for redundant variables~\citep{fop_variable_2018}.

Feature selection should not be mistaken with dimensionality reduction, sometimes called feature reduction, which is the process of finding a latent space of lower dimension leveraging good manifolds for clustering, e.g. using matrix factorisation~\citep{shen_integrative_2012}. In this sense, methods seeking a sparse subspace for spectral clustering~\citep{peng_feature_2016} or for KMeans clustering through PCA~\citep{long_flexible_2021} are discriminative. However, the nature of such projection forces the clustering to be done according to linear boundaries due to the projections. Still, by enforcing the projection matrix to be sparse, feature selection can be recovered in the original space~\citep{bouveyron_discriminative_2014}. Similarly, subspace clustering seeks to find clusters in different subspaces of the data~\citep{zografos_discriminative_2013, chen_discriminative_2018} and is thus an extension of feature selection~\citep{parsons_subspace_2004}, particularly with the motivation that several latent variables could explain the heterogeneity of the data~\citep{vandewalle_multi-partitions_2020}. However, such problems usually incorporate a mechanism to merge clusters, which is challenging as well, while we are interested in a method that selects features while producing a single clustering output.

Finally, clustering models in feature selection are often model-based~\citep{scrucca_clustvarsel_2018, raftery_variable_2006, maugis_variable_2009}, which implies that they assume a parametric mixture model that can explain the distribution of the data, including the distribution of irrelevant variables. To perform well, these methods need a good selection criterion to compare models with one another~\citep{raftery_variable_2006, marbac_variable_2020, maugis_variable_2009}. To the best of our knowledge, there do not exist models for joint feature selection and clustering in the discriminative sense of~\citet{minka_discriminative_2005} and ~\citet{krause_discriminative_2010}, i.e., models that only design $p_\theta(y|\vec{x})$ with end-to-end training. Finally, most of these generative wrapper methods hardly scale in both sample quantity and/or variable quantity.

%%%%%%%%%%%%%%%%%%%%%%%%%%%%%%%%%%%%%%%%%%%%%%%%%%%%%%%%%%%%%%%%%%%%%%%%%
% THE SPARSE GEMINI
%%%%%%%%%%%%%%%%%%%%%%%%%%%%%%%%%%%%%%%%%%%%%%%%%%%%%%%%%%%%%%%%%%%%%%%%%

\section{The Sparse GEMINI}
\label{sec:method}

Sparse GEMINI is a combination of the generalised mutual information objective for discriminative clustering~\citep{ohl_generalised_2023} with the LassoNet framework for feature selection~\citep{lemhadri_lassonet_2021} in neural networks, including the sparse logistic regression as a specific case. The model is summarised in Figure~\ref{fig:sparse_gemini_overview}.

\subsection{The GEMINI objective}

Let $\mathcal{D} = \{\vec{x}_{i}\}_{i=1}^N \subset \mathcal{X}$ a dataset of $N$ observations of dimension $d$. We note each feature $\vec{x}_{ij} \in \mathcal{X}_j$, thus: $\mathcal{X} = \prod_{j=1}^d \mathcal{X}_j$. We seek to cluster this dataset by learning a distribution $p_\theta(y|\vec{x})$ where $y$ is a discrete variable taking $K$ values.  This distribution is defined by a softmax-ended function which ensures that the elements of the resulting vector add up to 1:

\begin{equation}
    y|\vec{x} \sim \text{Categorical}(\text{SoftMax} \circ f_\theta(\vec{x})),
\end{equation}

\noindent
where $f_\theta: \mathcal{X} \mapsto \mathbb{R}^K$ has parameters $\theta$. For example, setting $f$ to an affine function recovers the multiclass logistic regression. In order to perform clustering with $f$ as a discriminative distribution, we train the parameters $\theta$ using a generalised mutual information (GEMINI, ~\citealp{ohl_generalised_2023}). This objective was introduced to circumvent the need for parametric assumptions regarding $p(\vec{x})$ in clustering and thus leads to designing only a discriminative clustering model $p_\theta(y|\vec{x})$~\citep{minka_discriminative_2005}. With the help of Bayes theorem, this objective can be estimated without assumptions of the data distribution $p(\vec{x})$ using only the output of the clustering distribution $p_\theta(y|\vec{x})$. Note that, as we avoid parametric hypotheses on the data distribution, we do not have a relationship between the data and the parameters, hence the writing $p(\vec{x})$ without $\theta$. Despite the absence of assumptions, we are able to sample from $p(\vec{x})$ owing to the dataset we have at hand. Overall, the GEMINI aims at separating according to a distance $D$ the cluster distributions from either the data distribution (one-vs-all):

\begin{equation}\label{eq:gemini_ova}
    \I_D^\text{ova}(\vec{x};y|\theta) = \E_{y \sim p_\theta(y)} \left[ D(p_\theta(\vec{x}|y)\|p(\vec{x}))\right],
\end{equation}

\noindent
or other cluster distributions (one-vs-one):

\begin{equation}\label{eq:gemini_ovo}
    \I_D^\text{ovo}(\vec{x};y|\theta) = \E_{y_1, y_2 \sim p_\theta(y)} \left[ D(p_\theta(\vec{x}|y_1)\|p(\vec{x}|y_2))\right].
\end{equation}

The novelty of GEMINI is to consider different types of distances $D$ between distributions with a special focus on the maximum mean discrepancy (MMD, \citealp{gretton_kernel_2012}) or the Wasserstein distance~\citep{peyre_computational_2019} compared to former discriminative approaches using the standard mutual information~\citep{bridle_unsupervised_1992, krause_discriminative_2010} which require regularisation to learn. The former corresponds to the distance between the expectations of the respective distributions projected into a Hilbert space and the latter is an optimal transport distance describing the minimum of energy necessary to reshape one distribution as the other. Both of them incorporate geometrical information on the data respectively through a kernel $\kappa$ or a distance $\delta$ in the data space. Any neural network that is trainable through cross-entropy loss can be switched to unsupervised learning at the cost of choosing a metric or kernel in the data space. While we cannot compute the true GEMINI values, we can estimate them using Bayes theorem to get an expression depending only on the predictions of the model. For instance, the Wasserstein GEMINI can be estimated using importance weights to estimate the densities of each cluster distribution. The MMD can be computed with sums of kernel terms weighted by the predictions of the model~\citep[Table 1]{ohl_generalised_2023}. GEMINI can therefore train any discriminative model of the form $p_\theta(y|\vec{x})$ as long as the distance $D$ can be evaluated using only the model's outputs. Thus, the written GEMINI objective becomes for a batch of size $N$:

\begin{equation}
    \hat{\I_D} = \sum_{k=1}^K f_D (\{p_\theta(y=k|\vec{x}=\vec{x}_1), \ldots, p_\theta(y=k|\vec{x}=\vec{x}_N)\}),
\end{equation}

\noindent where $f_D$ is a function depending on the chosen distance. For instance, if $D$ is the KL distance, we can show that:

\begin{multline}
    f_D(\{p_\theta(y=k|\vec{x}=\vec{x}_1), \ldots, p_\theta(y=k|\vec{x}=\vec{x}_N)\}) = \\ \sum_{i=1}^N p_\theta(y=k|\vec{x}=\vec{x}_i) \log \frac{p_\theta(y=k|\vec{x}=\vec{x}_i)}{p_\theta(y=k)}.
\end{multline}

All GEMINIs from~\citet{ohl_generalised_2022} are available in the GemClus package that we will present in Section~\ref{sec:package}.

\subsection{Sparse models}
\subsubsection{Unsupervised logistic regression architecture}
We start with the simplest discriminative model for variable selection: logistic regression. This corresponds to the case where our distribution $p_\theta(y|\vec{x})$ is characterised by the set of linear functions:

\begin{equation}
    \mathcal{F} = \{f_\theta: \vec{x} \mapsto \theta^\top\vec{x}\},
\end{equation}

\noindent
with $\theta \in \mathbb{R}^{d\times K}$ for $d$ features and $K$ clusters. Notice the absence of bias as this linear model is a sub-case of the neural network model covered in the next section. To properly ensure that vector weights are eliminated at once, a group-lasso penalty is preferred~\citep[Section 4.3]{hastie_statistical_2015} also known as $\ell_1/\ell_2$ penalty~\citep{bach_optimization_2012}. We consider a user-defined partition of the input features into $G \leq d$ groups, each with associated parameter subset $\theta_j$. Note that the dimensions of $\theta_j$ vary depending on the number of features within the $j$-th group. For example, a categorical variable taking $M$ values transformed into a one-hot-encoded vector of dimension $M$ can be associated to a single group. Note that all clusters use the same subset of selected groups of variables. Thus, the optimal parameters should satisfy:

\begin{equation}
    \hat{\theta} = \argmax_\theta \I_D(\vec{x};y|\theta) - \lambda \sum_{j=1}^G \|\theta_j\|_2.
\end{equation}

This is exactly the same objective formulation as the supervised multi-class Lasso if we replace the GEMINI by the maximum likelihood or any other supervised loss. Notice that $\lambda$ is positive because we seek to simultaneously maximise the GEMINI and minimise the $\ell_1/\ell_2$ penalty. During training, the sparse linear parameter will progressively remove variables by setting all grouped parameters to 0. If we set $G=d$, then each variable can be removed on its own as there are no groups of variables. A similar objective without group-lasso and using the standard mutual information can be found in~\citep[Eq. 4]{youyong_discriminative_2014}, although specific initialisation strategies were required to circumvent the unspecificity of mutual information local maxima, as described by~\citet{ohl_generalised_2023}.

\subsubsection{The LassoNet architecture}

We extend this procedure to neural network by adapting the LassoNet~\citep{lemhadri_lassonet_2021} framework with GEMINIs. The neural network $f_\theta:\mathcal{X}\mapsto \mathbb{R}^K$ is taken from a family of architectures $\mathcal{F}$ consisting of one multi-layered perceptron (MLP) and a linear skip connection:

\begin{equation}
    \mathcal{F} = \{f_\theta: \vec{x} \mapsto g_{\vec{\omega}}(\vec{x}) + \vec{W}^\top\vec{x}\},
\end{equation}

\noindent
with $\theta=\{\vec{\omega},\vec{W}\}$ including $\vec{\omega}$ the parameters of the MLP $g_{\vec{\omega}}$ and $\vec{W}\in\mathbb{R}^{d\times K}$ the weights of a linear skip connection penalised by group-lasso. This leads to the same optimisation objective as previously with a focus on the skip connection parameters:

\begin{equation}
    \hat{\theta} = \argmax_\theta \I_D(\vec{x};y|\theta) - \lambda \sum_{j=1}^G\|\vec{W}_j\|_2,
\end{equation}

\noindent
with $\vec{W}_j$, the weights of the $j$-th group of features from $\vec{W}$. It is a matrix of dimension $K$ times the group size. As the sparse skip connection $\vec{W}$ loses some feature subset, we must force the MLP to drop this same subset of features as well. Therefore the weights of the first layer $\vec{\omega}^{(1)}$ are constrained such that:

\begin{equation}
    \|\vec{\omega}^{(1)}_j\|_\infty \leq M\|\vec{W}_{j}\|_2, \forall j \leq G.
\end{equation}

\noindent
where $M$ is called the hierarchy coefficient. Thus, when a feature $j$ is eliminated, all weights starting from this feature in the MLP will be equal to 0 as well. When $M=0$, the method is equivalent to the penalised logistic regression from the previous section because all entry weights of the MLP are equal to zero, hence passing no information. 

Interestingly, while the constraints are designed to specifically select features, dimension reduction can be performed as well by extracting representations from lower-dimension layers in the network $g_{\vec{\omega}}$. However, this intermediate representation would not be complete as it misses the information from the skip connection.

\section{Optimisation}
\label{sec:optimisation}

\subsection{Training and model selection}
\label{ssec:selection_process}

We follow~\citet{lemhadri_lassonet_2021} in proposing a \emph{dense-to-sparse} training strategy for the penalty coefficient. Training is carried along a path where the $\ell_1$ penalty parameter $\lambda$ is geometrically increased: $\lambda = \lambda_0 \rho^t$ ($\rho >1$) at time step $t$ after an initial step without $\ell_1$ penalty. We stop when the number of remaining features used by the model is below a user-defined threshold $0 < F_\text{thres} <d$  which can be thought of as the minimum number of useful variables required. Each time the number of features decreases during training, we save its associate intermediate model. It is thus possible to restore any intermediate model to get a clustering on different subsets of variables.

Once the training is finished, we look again at all GEMINI scores during the feature decrease and select the model with the minimum of features that managed to remain in the arbitrary range of 90\% of the best GEMINI value. This best value is most of the time the loss evaluated with the model exploiting all features. Still, automatic model selection may not be necessary and in this case, we can look at all intermediate models produced during the path as a set of possible solutions for clustering with different number of selected variables.

As GEMINI depends on a metric defined in the data space, the metric will still be computed using all features despite selection by the model. Consequently, we propose a less grounded yet efficient training mode in appendix~\ref{app:dynamic_training} taking into account variable selection in the metric computation. However, this training mode yields incomparable GEMINI scores due to the change of metric definition, rendering the selection strategy inapplicable.

Interestingly, as the MMD-GEMINI maximisation is equivalent to a kernel KMeans objective, as shown by~\citet{franca_kernel_2020}, a subset of the method resembles to a sparse \emph{kernel} KMeans. However, unlike related works~\citep{franca_kernel_2020, witten_framework_2010}, the selection process is done through gradient descent, i.e. without indicative variables, and without the definition of explicit centroids, thus being less strict regarding the number of clusters to find, i.e. GEMINI models can find fewer clusters than asked.

\subsection{Gradient considerations}

We offer here more detailed insights on the gradients of the model for ensuring true variable elimination and how the GEMINI behaves.

\subsubsection{Proximal gradients}

To ensure the convergence of the parameters to 0 upon elimination, we adopt a proximal gradient strategy. In the case of sparse logistic regression, the gradient \emph{ascent} hence follows the two classical steps:

\begin{equation}
    \beta = \theta^t + \eta^t\nabla_\theta \I_D(\vec{x};y|\theta),
\end{equation}
\begin{equation}
    \theta^{t+1}_j = \mathcal{S}_{\rho^t\lambda \|\beta_j\|_2}(\beta_j), \forall j\leq d,
\end{equation}

\noindent
where $\eta^t$ is the learning at timestep $t$. The soft-thresholding operation $\mathcal{S}_\alpha$:

\begin{equation}
    \mathcal{S}_\alpha(x) = \text{sign}(x)\max\{0, |x|-\alpha\},
\end{equation}

\noindent
is the closed-form solution of the proximal operator to project the parameters on the constrained space due to the group-lasso penalty~\citep[Section 5.3.3]{hastie_statistical_2015}. Consequently, we are sure to obtain true zeroes in the linear weights of the logistic regression or the weights of the skip connection for the neural network. For the case of the complete neural network model, \citet{lemhadri_lassonet_2021} gracefully provide a proximal gradient operation to satisfy inequality constraints during training time which guarantees true zeros in the first MLP layer as well.

\subsubsection{GEMINI gradients}

We extend our initial work on gradients~\citep{ohl_generalised_2023} here by providing explicit gradients for the GEMINI functions and some explanations. Let us consider for example the derivative of the OvA MMD-GEMINI with respect to the probability output of some $p_\theta(y=k|\vec{x}=\vec{x}_i)$ for a sample $\vec{x}_i$ in a batch containing $N$ samples. Adapting Eq.~(\ref{eq:gemini_ova}) to this specific distance, the objective to maximise is:

\begin{equation}
    \mathcal{I} = \mathbb{E}_{y \sim p_\theta(y)} \left[ \text{MMD}(p_\theta(\vec{x}|y) \| p_\text{data}(\vec{x}))\right].
\end{equation}

We can show that the gradient of an estimated OvA MMD-GEMINI $\hat{\I}$ is:

\begin{multline}
    \diff{\hat{\I}}{p_\theta(y=k|\vec{x}=\vec{x}_i)} = \frac{1}{\text{MMD}_k} \left[\frac{1}{N^3}\sum_{j,l}^N\kappa(\vec{x}_j,\vec{x}_l) \right.\\-\frac{2}{N^3}\sum_{j}^N\sum_{l=1}^N\kappa(\vec{x}_j,\vec{x}_l)\frac{p_\theta(y=k|\vec{x}_l)}{p_\theta(y=k)} \\\left.+ \sum_{j}^N \kappa(\vec{x}_i,\vec{x}_j)\left(\frac{p_\theta(y=k|\vec{x}=\vec{x}_j)}{p_\theta(y=k)}-1\right)\right],
\end{multline}

\noindent
where $\text{MMD}_k$ stands for the MMD between the $k$-th cluster and the data distribution and $\kappa$ is the kernel of choice. Regarding the interpretation of the gradient, we can observe that for all clusters $k$, the gradient is divided by the respective MMD distance. In other words, the smaller the MMD the stronger the gradient: we need to increase the MMD. All clusters $k$ get their own common gradient throughout samples. This is due to the interdependence nature of the OvA MMD that requires batches of samples to be evaluated and cannot be performed on one sample at a time. Moreover, this common term comprises a constant which is the estimated participation of the data to the MMD. The second term is the cross-contribution of $p_\theta(y|\vec{x})$ and $p(\vec{x})$ in the evaluation of the distance. The stronger this cross-contribution is (and so the smaller the MMD), the smaller the gradient. Finally, each individual sample receives finally its own gradient which is its weighted average kernel strength by $\frac{p_\theta(y=k|\vec{x})}{p(y=k)}-1$. Thus, the greater the conditional distribution of a sample compared to the cluster distribution, the stronger the gradient. Indeed, the more representative a sample of the cluster distribution, the greater its contribution to pulling the gradient to this representation.

We give further details on other gradients as well in Appendix~\ref{app:mmd_ova_computations},\ref{app:mmd_ovo_computations} and \ref{app:wasserstein_computations}. The latter is dedicated to the Wasserstein GEMINI.

\subsection{Implementations}
\label{sec:package}

% \begin{lstlisting}[float, language=PythonPlus, style=colorEX, caption=\centering An example of \emph{gemclus} loading and data clustering
% \label{list:gemclus_example}]
% from gemclus.sparse import SparseLinearMMD
% from sklearn.datasets import load_breast_cancer
% # load data
% X, _ = load_breast_cancer(return_X_y=True)
% # Create simple logistic regression model and do clustering
% y_pred = SparseLinearMMD(n_clusters=2).fit_predict(X)
% \end{lstlisting}
% \begin{lstlisting}[float, language=PythonPlus, style=colorEX,caption=\centering The package \emph{gemclus} incorporates as well the basic logistic regression with regularised mutual information by~\citet{krause_discriminative_2010}\label{list:rim_example}]
% from gemclus.linear import RIM
% y_pred = RIM(n_clusters=3).fit_predict(X)
% \end{lstlisting}

Owing to the exact computations of the gradients, we developed a Python package encompassing all sparse GEMINI methods as well as original non-sparse GEMINI methods~\citep{ohl_generalised_2023} named \emph{GemClus}\footnote{The GemClus package can be found at \url{https://gemini-clustering.github.io/}}. Overall, the package is designed for small datasets as it encompasses the precise and optimised computations of the gradients with lightweight dependencies, rather than using the automatic differentiation processes of heavy packages like PyTorch. The code of this paper was written using the latter\footnote{The code for this paper can be found at \url{https://github.com/oshillou/SparseGEMINI}.}.
Results may consequently slightly differ. We give detailed examples of code snippets in App.~\ref{app:gemclus_details}, including how to reproduce the numerical experiments from Section~\ref{ssec:numerical_experiments}. %We give an example of code in Listing~\ref{list:gemclus_example}.

%We also want to extend this package to other discriminative clustering methods for potentially small-scale datasets. Therefore, we include an implementation of the regularised mutual information (RIM) model by \citet{krause_discriminative_2010} as shown in Listing~\ref{list:rim_example} because we consider this model to be one of the very first proposed in the domain yet find few satisfying implementations. 

%%%%%%%%%%%%%%%%%%%%%%%%%%%%%%%%%%%%%%%%%%%%%%%%%%%%%%%%%%%%%%%%%%%%%%%%%
% EXPERIMENTS
%%%%%%%%%%%%%%%%%%%%%%%%%%%%%%%%%%%%%%%%%%%%%%%%%%%%%%%%%%%%%%%%%%%%%%%%%

\section{Experiments}
\label{sec:experiments}
A brief summary of the datasets used in these experiments can be found in table~\ref{tab:dataset_description}.

\begin{table}[t]
    \centering
    \caption{Brief description of datasets involved in experiments}
    \label{tab:dataset_description}
    \vskip 0.1in
    \begin{tabular}{c c c c}
        \toprule
        Name& Samples&Features & \#Classes\\
        \midrule
        US-Congress&435&16&2\\
        Heart-statlog&270&13&2\\
        MNIST&12000&784&10\\
        MNIST-BR&12000&784&10\\
        Prostate-BCR&171&25904&2\\
        \bottomrule
    \end{tabular}
\end{table}

\subsection{Metrics}

Depending on the experiments for comparison purposes, we report 3 different metrics. The adjusted rand index (ARI,~\citealp{hubert_comparing_1985}) describes how close the clustering is to the classes, with a correction to random guesses. The variable selection error rate (VSER), for instance used by~\citet{celeux_comparing_2014}, describes the percentage of variables that the model erroneously omitted or accepted, therefore the lower the better. We finally report the correct variable rate (CVR) which describes how many of the expected variables were selected: higher is better. For example, a model selecting all variables of a dataset with $d$ variables and $d^\prime$ good variables will get a CVR of 100\% and a VSER of $1-\frac{d^\prime}{d}$.

\subsection{Default hyperparameters}

We set the hierarchy coefficient to $M=10$, as \citet{lemhadri_lassonet_2021} report that this value seems to ``work well for a variety of datasets''. We also report the performances for the logistic regression mode when $M=0$. The optimiser for the initial training step with $\lambda=0$ is Adam~\citep{kingma_adam_2014} with a learning rate of $10^{-3}$ while other steps are done with SGD with momentum 0.9 and the same learning rate. Most of our experiments are done with 100 epochs per step with early stopping as soon as the global objective does not improve by 1\% for 10 consecutive epochs. The early stopping criterion is evaluated on the same training set since we do not seek to separate the dataset in train and validation sets in clustering. All activation functions are ReLUs. The default starting penalty is $\lambda_0=1$ with a 5\% increase per step. We keep the linear kernel and the Euclidean distance respectively in conjunction with the MMD and Wasserstein distances when evaluating the GEMINI. Finally, we evaluate in most experiments the method with the exact same number of clusters as the number of known (supervised) labels.

\subsection{Numerical experiments}
\label{ssec:numerical_experiments}

\begin{sidewaystable*}%[!hbtp]
\centering
\caption{Performances of Sparse GEMINI (OvO only) on synthetic datasets after 20 runs. We compare our performances against other methods. S stands for a scenario of the first synthetic dataset and D2 stands for the second synthetic dataset. Standard deviation is reported in subscript.}\label{tab:synthetic_datasets_results}
    \vskip 0.1in
\subfloat[ARI scores (greater is better)] {
	\begin{tabular}{c c c c c c c c c}
	\toprule
	&\multirow{2}{*}{Sparse KMeans}&\multirow{2}{*}{ClustVarSel}&\multirow{2}{*}{VSCC}&\multirow{2}{*}{SFEM}&\multicolumn{2}{c}{MMD-GEMINI}&\multicolumn{2}{c}{Wasserstein-GEMINI}\\
	\cmidrule{6-7}\cmidrule{8-9}
	&&&&&Logistic&MLP&Logistic&MLP\\
	\midrule
	S1&0.09\std{0.08}&0.05\std{0.07}&0.00\std{0.02}&\textbf{0.13\std{0.11}}&\textbf{0.14\std{0.11}}&0.08\std{0.07}&0.09\std{0.12}&0.04\std{0.07}\\
	S2&\textbf{0.80\std{0.15}}&0.20\std{0.22}&0.02\std{0.03}&0.72\std{0.17}&0.59\std{0.17}&0.52\std{0.23}&0.44\std{0.18}&0.43\std{0.15}\\
	S3&0.11\std{0.03}&0.04\std{0.10}&0.15\std{0.10}&\textbf{0.22\std{0.05}}&\textbf{0.21\std{0.05}}&\textbf{0.21\std{0.04}}&0.14\std{0.05}&0.10\std{0.05}\\
	S4&\textbf{0.87\std{0.04}}&\textbf{0.88\std{0.04}}&0.84\std{0.12}&\textbf{0.87\std{0.04}}&0.74\std{0.07}&\textbf{0.86\std{0.05}}&0.73\std{0.19}&0.82\std{0.11}\\
	S5&\textbf{0.87\std{0.03}}&0.65\std{0.38}&0.00\std{0.00}&0.83\std{0.03}&0.76\std{0.05}&\textbf{0.86\std{0.03}}&0.59\std{0.20}&0.67\std{0.20}\\
	\cmidrule{2-9}
	D2&0.31\std{0.03}&\textbf{0.60\std{0.02}}&0.58\std{0.02}&0.58\std{0.01}&0.57\std{0.01}&0.54\std{0.03}&0.57\std{0.01}&0.55\std{0.02}\\
	\bottomrule
	\end{tabular}
}\hfill
\subfloat[VSER scores (lower is better)]{
	\begin{tabular}{c c c c c c c c c}
	\toprule
	&\multirow{2}{*}{Sparse KMeans}&\multirow{2}{*}{ClustVarSel}&\multirow{2}{*}{VSCC}&\multirow{2}{*}{SFEM}&\multicolumn{2}{c}{MMD-GEMINI}&\multicolumn{2}{c}{Wasserstein-GEMINI}\\
	\cmidrule{6-7}\cmidrule{8-9}
	&&&&&Logistic&MLP&Logistic&MLP\\
	\midrule
	S1&0.31\std{0.22}&0.28\std{0.06}&0.72\std{0.15}&\textbf{0.24\std{0.07}}&0.44\std{0.15}&0.43\std{0.11}&0.51\std{0.11}&0.56\std{0.13}\\
    S2&0.75\std{0.18}&0.29\std{0.07}&0.73\std{0.09}&0.27\std{0.08}&\textbf{0.06\std{0.05}}&0.11\std{0.07}&0.15\std{0.10}&0.24\std{0.12}\\
    S3&0.47\std{0.34}&0.25\std{0.07}&0.65\std{0.22}&0.20\std{0.04}&\textbf{0.07\std{0.05}}&0.20\std{0.11}&0.20\std{0.12}&0.56\std{0.14}\\
    S4&0.80\std{0.00}&\textbf{0.01\std{0.03}}&0.64\std{0.29}&0.23\std{0.08}&\textbf{0.00\std{0.00}}&\textbf{0.00\std{0.00}}&\textbf{0.01\std{0.03}}&\textbf{0.00\std{0.02}}\\
    S5&0.95\std{0.00}&0.05\std{0.03}&0.95\std{0.00}&0.10\std{0.02}&\textbf{0.00\std{0.00}}&\textbf{0.00\std{0.00}}&\textbf{0.01\std{0.01}}&\textbf{0.01\std{0.01}}\\
	\cmidrule{2-9}
	D2&0.84\std{0.06}&0.00\std{0.00}&0.74\std{0.13}&0.52\std{0.08}&0.29\std{0.00}&0.31\std{0.04}&0.29\std{0.00}&0.29\std{0.00}\\
	\bottomrule
	\end{tabular}
}\hfill
\subfloat[CVR scores (greater is better)]{
	\begin{tabular}{c c c c c c c c c}
	\toprule
	&\multirow{2}{*}{Sparse KMeans}&\multirow{2}{*}{ClustVarSel}&\multirow{2}{*}{VSCC}&\multirow{2}{*}{SFEM}&\multicolumn{2}{c}{MMD-GEMINI}&\multicolumn{2}{c}{Wasserstein-GEMINI}\\
	\cmidrule{6-7}\cmidrule{8-9}
	&&&&&Logistic&MLP&Logistic&MLP\\
	\midrule
	S1&0.53\std{0.31}&0.11\std{0.10}&\textbf{0.87\std{0.23}}&0.28\std{0.18}&0.60\std{0.26}&0.64\std{0.19}&0.63\std{0.23}&0.59\std{0.17}\\
    S2&\textbf{1.00\std{0.00}}&0.14\std{0.17}&0.66\std{0.39}&0.39\std{0.17}&0.93\std{0.10}&0.82\std{0.22}&0.83\std{0.13}&0.78\std{0.14}\\
	S3&\textbf{1.00\std{0.00}}&0.19\std{0.30}&0.97\std{0.13}&0.27\std{0.15}&0.95\std{0.09}&0.99\std{0.04}&0.68\std{0.25}&0.92\std{0.12}\\
	S4&\textbf{1.00\std{0.00}}&0.19\std{0.30}&0.97\std{0.13}&0.27\std{0.15}&1.00\std{0.00}&1.00\std{0.00}&0.99\std{0.04}&0.99\std{0.04}\\
	S5&\textbf{1.00\std{0.00}}&0.75\std{0.44}&\textbf{1.00\std{0.00}}&0.66\std{0.18}&\textbf{1.00\std{0.00}}&\textbf{1.00\std{0.00}}&0.94\std{0.11}&0.96\std{0.10}\\
	\cmidrule{2-9}
	D2&0.98\std{0.11}&\textbf{1.00\std{0.00}}&\textbf{1.00\std{0.00}}&\textbf{1.00\std{0.00}}&0.00\std{0.00}&0.00\std{0.00}&0.00\std{0.00}&0.00\std{0.00}\\
	\bottomrule
	\end{tabular}
}
\end{sidewaystable*}

\begin{table*}
    \centering
    \caption{Average regret scores (lower is better) between Sparse GEMINI for OvO MMD GEMINI against the best performing method per dataset.}
    \label{tab:regret_score}
    \begin{tabular}{ccccccc}
        \toprule
        Method & \multicolumn{2}{c}{ARI} & \multicolumn{2}{c}{VSER} & \multicolumn{2}{c}{CVR} \\
        \cmidrule(r){2-3}\cmidrule(lr){4-5}\cmidrule(l){6-7}
        Dataset &  Linear &   MLP & Linear & MLP & Linear & MLP \\
        \midrule
        1s1     &  0.00\std{0.11} &  0.06\std{0.07} &  0.17\std{0.15} &  0.16\std{0.11} &  0.28\std{0.26} &   0.24\std{0.19} \\
        1s2     &  0.19\std{0.17} &  0.26\std{0.23} &  0.02\std{0.05} &  0.07\std{0.07} &  0.07\std{0.10} &   0.18\std{0.22} \\
        1s3     &  0.02\std{0.05} &  0.02\std{0.04} &  0.01\std{0.05} &  0.14\std{0.11} &  0.04\std{0.09} &  0.00\std{0.04} \\
        1s4     &  0.14\std{0.07} &  0.02\std{0.05} &  0.00\std{0.00} &  0.00\std{0.00} &  0.00\std{0.00} &   0.00\std{0.00} \\
        1s5     &  0.12\std{0.05} &  0.02\std{0.03} &  0.00\std{0.00} &  0.00\std{0.00} &  0.00\std{0.00} &   0.00\std{0.00} \\
        \cmidrule{2-7}
        2s2     &  0.02\std{0.01} &  0.05\std{0.03} &  0.29\std{0.00} &  0.31\std{0.04} &  1.00\std{0.00} &   1.00\std{0.00} \\
        \bottomrule
    \end{tabular}
\end{table*}

% \begin{lstlisting}[float,language=PythonPlus, style=colorEX, caption=\centering An example of sparse GEMINI model fitting the 5th scenario of the synthetic datasets\label{list:code_experiment}]
% from gemclus.sparse import SparseMLPMMD
% from gemclus.data import celeux_one

% # Generate the data according to the 5th scenario
% X,y = celeux_one(n=300, p=95, mu=1.7)
% # Prepare the model: MLP with the OvA MMD-GEMINI for 3 clusters
% model = SparseMLPMMD(n_clusters=3)
% # Progressively increase the penalty until all features are removed
% # res contains the history of feature selection and best model weights
% res = model.path(X)
% \end{lstlisting}

\begin{figure}
    \centering
    \subfloat[Dataset 1 Scenario 5]{
        \includegraphics[width=0.45\linewidth]{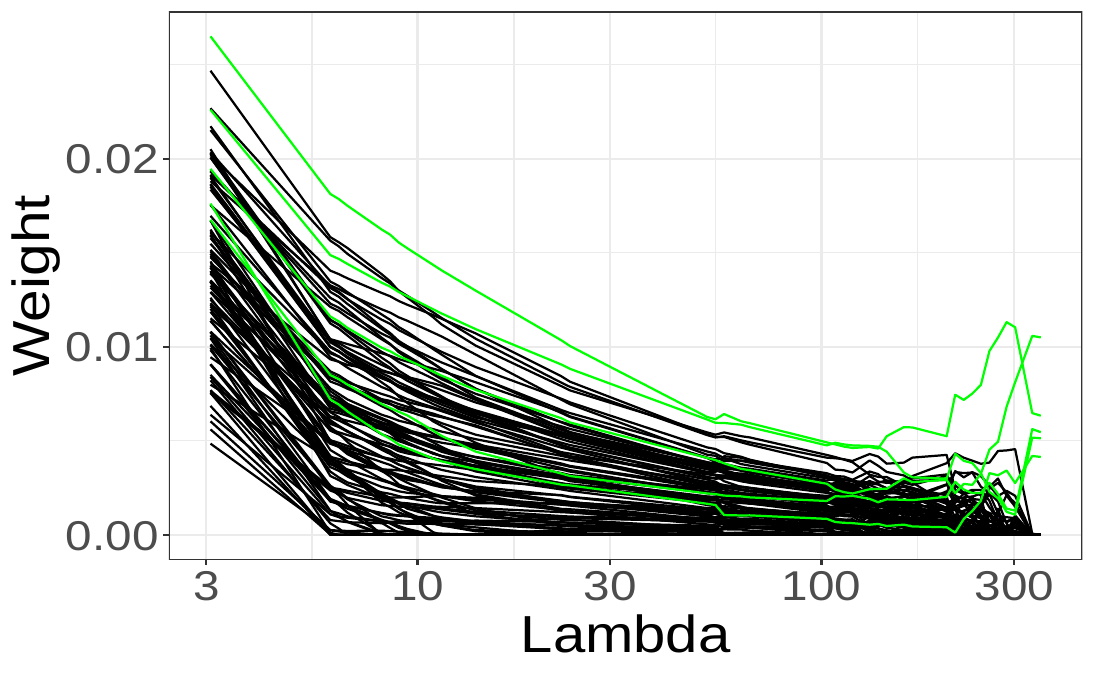}
    }
    \subfloat[Dataset 2]{
        \includegraphics[width=0.45\linewidth]{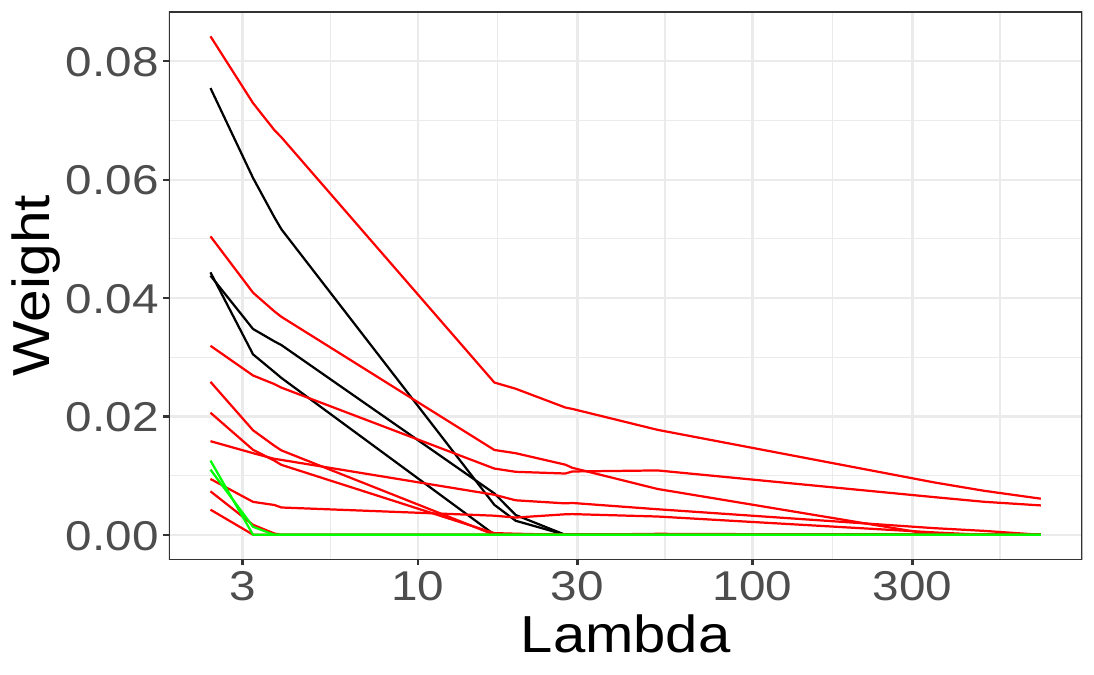}
    }
    \caption{Example of convergence of the norm of the weights of the skip connection for every feature during training for the OvA Wasserstein objective. Green lines are the informative variables, black lines are the noise and red are the correlated variables. (a) In the case of noisy variables, Sparse GEMINI can recover the informative variables. (b) In the presence of redundant variables, Sparse GEMINI eliminates informative variables to keep the redundant ones.}
    \label{fig:celeux_2_weights_history}
\end{figure}

We tested Sparse GEMINI on two synthetic datasets proposed by  \citet{celeux_comparing_2014} and also used by \citet{bouveyron_discriminative_2014} to first highlight some properties of the algorithm and compare it with competitors.% that we ran ourselves.

The first synthetic dataset consists of a few informative variables amidst noisy independent variables. The first 5 variables are informative and drawn from an equiprobable multivariate Gaussian mixture distribution of 3 components. All covariances are set to the identity matrix. The means are $\vec{\mu}_1=-\vec{\mu}_2 = \alpha \vec{1}$ and $\vec{\mu}_3 = \vec{0}$. All remaining $p$ variables follow independent noisy centred Gaussian distributions. The number of samples $N$, the mean proximity $\alpha$ and the number of non-informative variables $p$ vary over 5 scenarios. For the 2 first scenarios, we use $N=30$ samples and $N=300$ for others.  The scenarios 1 and 3 present the challenge of close Gaussian distributions with $\alpha=0.6$ while others use $\alpha=1.7$. Finally, we add $p=20$ noisy variables, except for the fifth scenario which takes up to $p=95$ uninformative variables. %This experiment can typically be done by running the algorithm from Listing~\ref{list:code_experiment} using \emph{gemclus}.

The second dataset consists of $n=2000$ samples of 14 variables, 2 of them informative and most others linearly dependent on the former. The Gaussian mixture is equiprobable with 4 Gaussian distributions of means $[0,0]$, $[4,0]$, $[0,2]$ and $[4,2]$ with identity covariances. The 9 following variables are sampled as follows:

\begin{multline}
    \vec{x}^{3-11} = [0,0,0.4,0.8,1.2,1.6,2.0,2.4,2.8]^\top \\+ {\vec{x}^{1-2}}^\top \left[\begin{array}{ccccccccc}0.5&2&0&-1&2&0.5&4&3&2\\1&0&3&2&-4&0&0.5&0&1\end{array}\right]+\vec{\epsilon},
\end{multline}

where $\vec{\epsilon} \sim \mathcal{N}(\vec{0},\vec{\Omega})$ with the covariance:

\begin{equation}
    \vec{\Omega} = \text{diag}\left(\vec{I}_3, 0.5\vec{I}_2, \text{diag}([1,3])\text{Rot}(\frac{\pi}{3}),\text{diag}[2,6]\text{Rot}(\frac{\pi}{6})\right).
\end{equation}

Finally, the last 3 variables are sampled independently from $\mathcal{N}([3.2,3.6,4],\vec{I}_3)$.

For all synthetic datasets, we asked training to stop with $F_\text{thres}$ set to the expected quantity of variables. We report the results of Sparse GEMINI in Table~\ref{tab:synthetic_datasets_results} after 20 runs. For detailed distributions' box plots, refer to Appendix~\ref{app:boxplots}. We compare our results against our own runs of other methods using their R package: SparseKMeans~\citep{witten_package_2013}, ClustVarSel~\citep{scrucca_clustvarsel_2018}, vscc~\citep{andrews_variable_2013, andrews_variable_2014} and SparseFisherEM~\citep{bouveyron_simultaneous_2012}. Due to the lack of space, we only report the scores for the one-vs-one GEMINI in Table~\ref{tab:synthetic_datasets_results}. Extensive results using the one-vs-all GEMINI with the two architectures can be found in Appendix~\ref{app:synthetic_wasserstein}.

It appears that the Sparse GEMINI is efficient in selecting the relevant variables when several others are noisy, especially with the OvO MMD objective while maintaining a high ARI. Moreover, while we do not systematically get the best ARI, our performances never fall far behind the most competitive method. We report in Table~\ref{tab:regret_score} the performances of the Sparse GEMINI MMD OvO that we deemed best against the best method for both architectures with regret scores. Regret scores are computed as the average difference between the scores of our model and the best performing one. Most scores are close to 0 for the third, fourth and fifth scenarios of the first dataset, except on the ARI. Our worst CVR is for the second dataset where we did not select the correct variables at all.  We can also observe in Table~\ref{tab:synthetic_datasets_results} that the MMD objective learns well despite the presence of few samples in scenarios 2 and 3 and that the usage of an MLP leads to a trade-off between ARI and VSER when we have enough samples. Additionally, the selection strategy often leads to selecting the correct number of variables for the MMD, except in scenarios 1 and 3 where the Gaussian distributions are close to each other which is hard given the large variance. For the Wasserstein objective, we notice that the performances in selection are improved with the presence of more samples. However, the clustering performances are worse than the MMD, which we can attribute to the contribution of the noisy variables to the computation of the distances between samples, thus troubling the holistic perspective of the Wasserstein distance on the cluster distribution. It also appears that we performed poorly at selecting the correct variables in the presence of redundancy in the second dataset. However, since all variables except 3 are correlated to the informative variables, we still managed to get a correct ARI on the dataset while using other variables. On average, the variables selected by our models were the 6th and the 8th variables. We focus on this difference of convergence in Figure~\ref{fig:celeux_2_weights_history} where we plot the norm of the skip connection per feature $\vec{W}_j$. In the case of noisy variables, we are able to recover them as the number of selected features decreases, whereas we eliminated the informative variable of the second dataset during the first steps. In general, Clustvarsel~\citep{scrucca_clustvarsel_2018} performed better on this type of synthetic dataset in terms of variable selection because it explicitly assumes a linear dependency between relevant variables and others.
%The obtained results are overall in line with the findings of~\citep{celeux_comparing_2014}. 
%We provide with Figure~\ref{fig:celeux_2_weights_history} an example of the evolution of the norm of the weights of the skip connection as the weight $\lambda$ increases.

\subsection{Examples on MNIST and variations}
\label{ssec:mnist_experiments}

\begin{figure}
    \centering
    \subfloat[][MNIST importance map]{
        \includegraphics[width=0.30\linewidth]{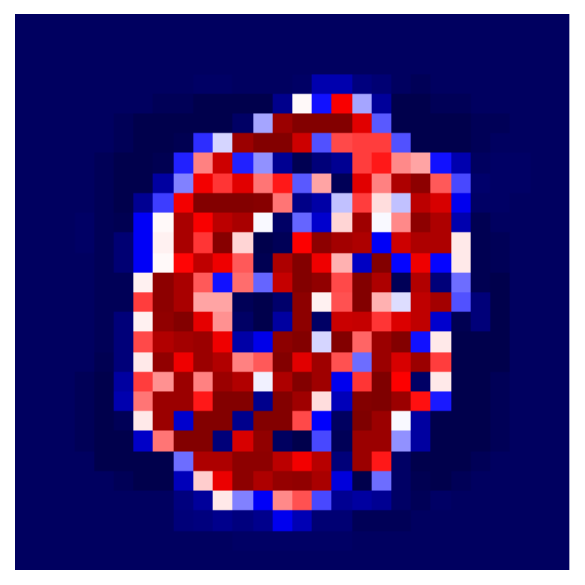}
        \label{sfig:mnist_features}
    }\hfil
    \subfloat[][MNIST GEMINI]{
        \includegraphics[width=0.55\linewidth]{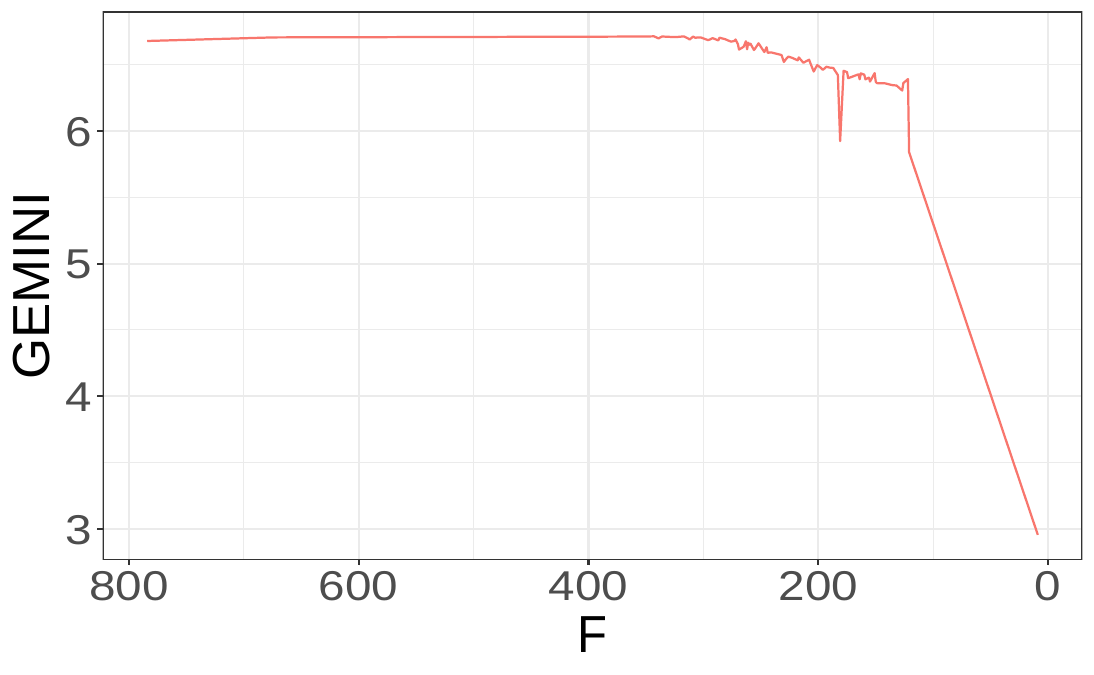}
        \label{sfig:mnist_gemini}
    }\\
    \subfloat[][MNIST-BR importance map]{
        \includegraphics[width=0.30\linewidth]{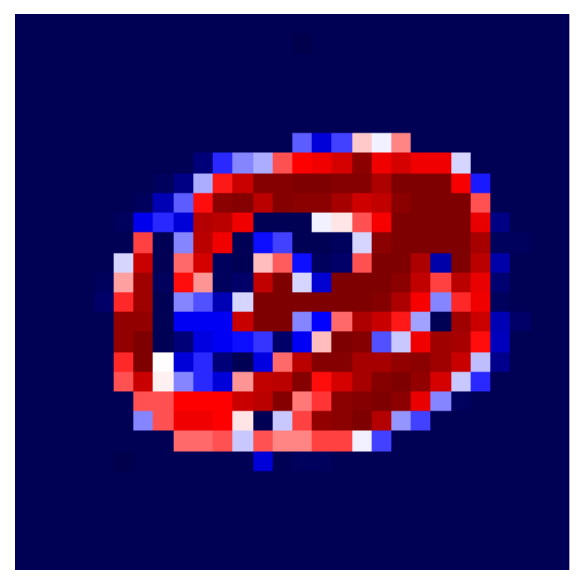}
        \label{sfig:mnist_br_features}
    }\hfil
    \subfloat[][MNIST-BR GEMINI]{
        \includegraphics[width=0.55\linewidth]{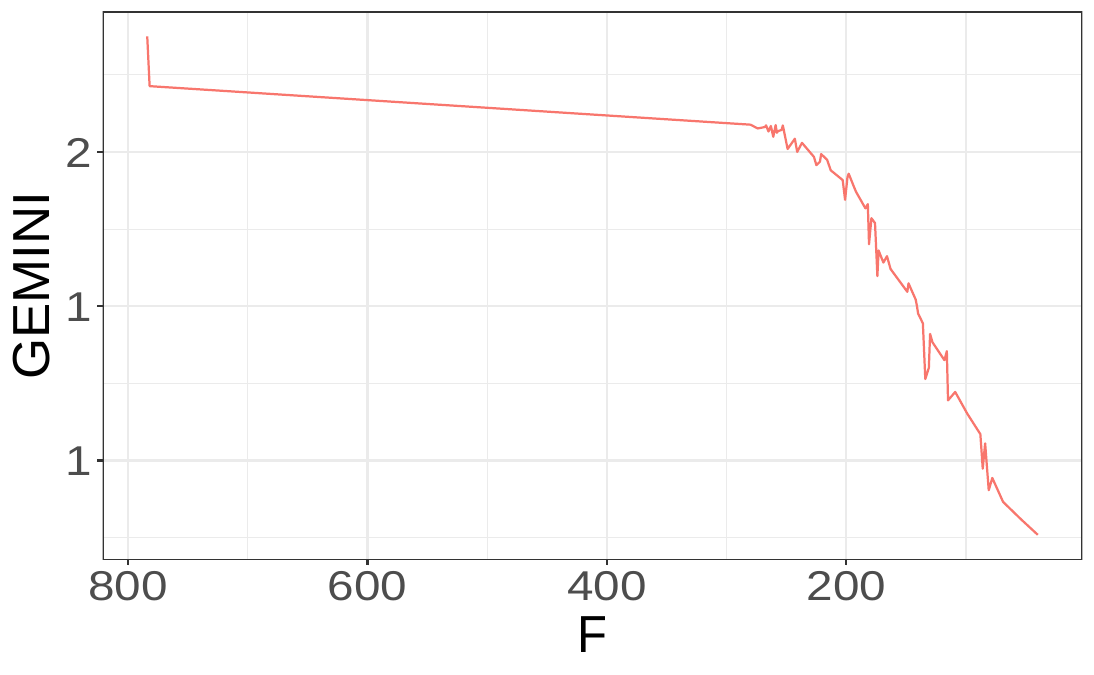}
        \label{sfig:mnist_br_gemini}
    }
    \caption{Relative importance of MNIST features after training of Sparse GEMINI with a log-scale color map. The blue features were eliminated at the first steps of $\lambda$, and the red features were eliminated last. On the right: evolution of the GEMINI depending on $\lambda$. $F$ stands for the number of selected features.}
    \label{fig:mnist_results}
\end{figure}

We also demonstrate performance of the Sparse GEMINI algorithm by running it on the MNIST dataset. The initial $\lambda_0$ was set to 40. Following~\citet{lemhadri_lassonet_2021}, we chose to stop training after finding 50 features. We also use 5\% of dropout inside an MLP with 2 hidden layers of 1200 dimensions each~\citep{hinton_improving_2012}. We report in Figure~\ref{fig:mnist_results} the selected features by the clustering algorithms and the evolution of the ARI. We also extended this experiment to the MNIST variations proposed by~\citet{larochelle_empirical_2007} showing the performance on the MNIST-BR dataset\footnote{Datasets were available at~\url{https://web.archive.org/web/20180519112150/http://www.iro.umontreal.ca/~lisa/twiki/bin/view.cgi/Public/MnistVariations}}, a challenging dataset for unsupervised variable selection~\citep{mattei_globally_2016}. This variation consists in samples of MNIST with the black background being replaced by uniform noise hence displaying conditional noise on the data. To be fair, we reduced MNIST to the first 12,000 samples of the training set in order to match the number of samples in MNIST-BR.

We observed in Figure~\ref{fig:mnist_results} that for both the default MNIST dataset and the MNIST-BR dataset despite the presence of noise, the feature map concentrates precisely on the good location of the digits in the picture. Following the GEMINI curves in the figures~\ref{sfig:mnist_gemini} and~\ref{sfig:mnist_br_gemini}, the respective selected numbers of features were 122 for MNIST and 243 for MNIST-BR. These chosen models also have a respective ARI of 0.34 for 7 clusters and 0.28 for 8 clusters. The presence of empty clusters is a possible outcome with GEMINI~\citep{ohl_generalised_2023} which contributed here to lowering the ARI when evaluating with the true digits targets.

%Moreover, both cases display a steady value of ARI despite the increase of $\lambda$, with a final collapse for the MNIST-BR because the last step of $\lambda$ was powerful enough to evince all features at once. Interestingly, the results on MNIST-BI exhibit an ARI of nearly 0 during training which is explained by the feature map. The model chose to base its clustering on the image backgrounds rather than the digits, making it thus likely unrelated to the labels matching the digits.

%We can immediately see that the importance map of Figure~\ref{fig:mnist_importance_map} is located where digits appear in the dataset. Moreover, the most important features are either located on the bottom left corner or on the vertical middle axis, which typically corresponds to regions where digits 0,2,3,5,8 and 9 are common, and for a distinction between 1, 4 and 7.

\subsection{Real datasets}
\label{ssec:real_datasets_experiments}

\subsubsection{OpenML datasets}
\label{sssec:openml_experiments}

\begin{figure}
    \centering
    \subfloat[][GEMINI Loss depending on selected feature number]{
        \includegraphics[width=0.45\linewidth]{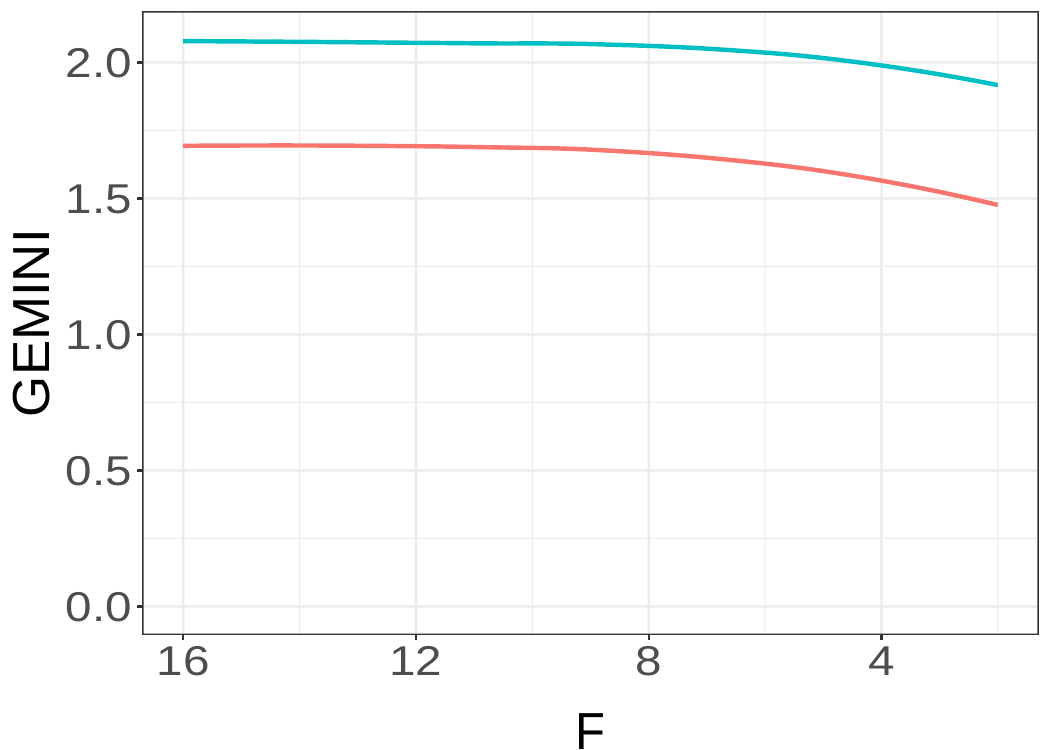}
        \label{sfig:congress_gemini}
    }\hfil
    \subfloat[][Selected features depending on $\lambda$]{
        \includegraphics[width=0.45\linewidth]{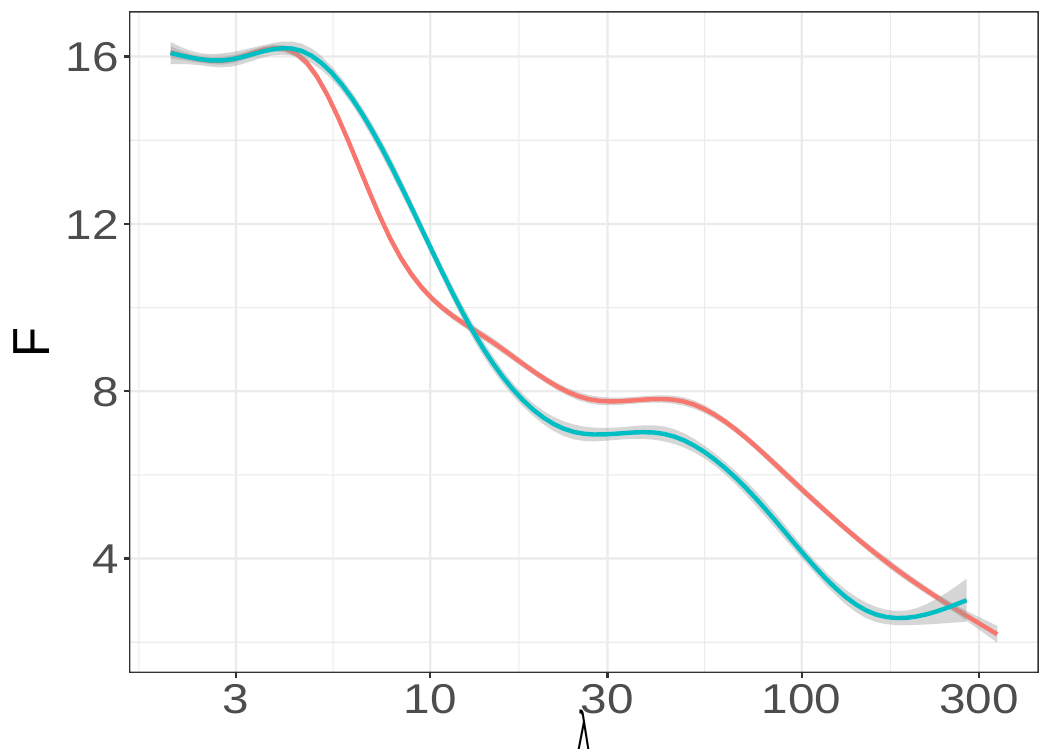}
        \label{sfig:congress_features}
    }
    \caption{Average training curves of Sparse GEMINI on the US Congress dataset over 50 runs. Blue lines are Wasserstein, red lines are MMD.}
    \label{fig:congress}
\end{figure}

\begin{figure}
    \centering
    \subfloat[][GEMINI depending on selected feature number]{
        \includegraphics[width=0.45\linewidth]{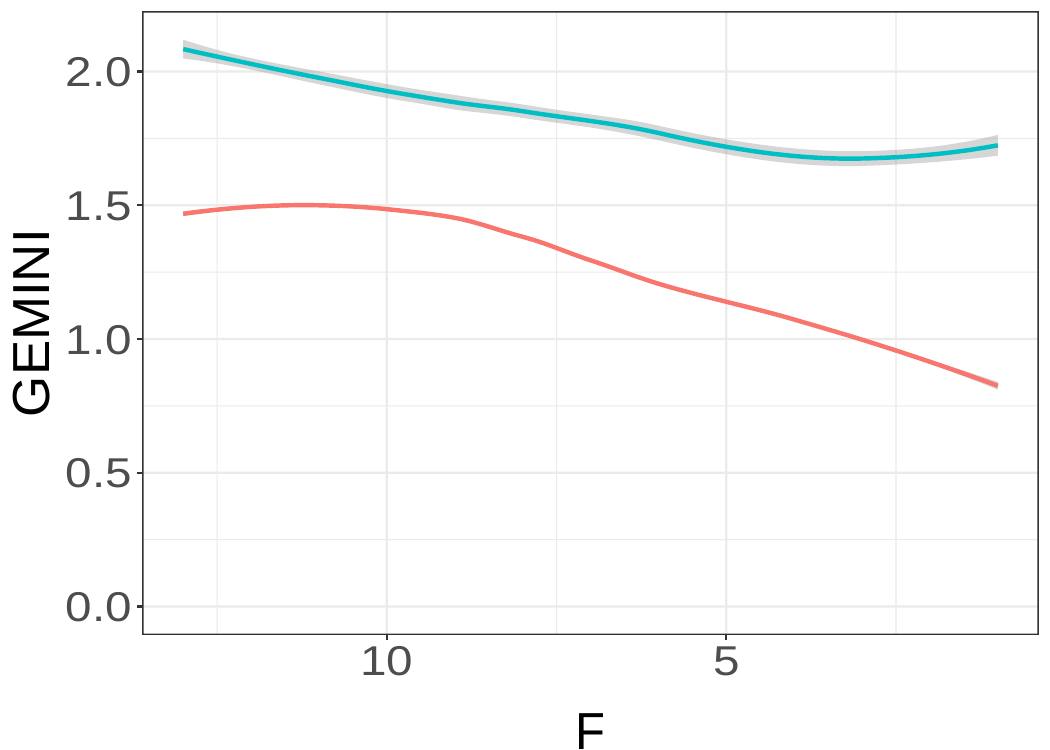}
        \label{sfig:heart_statlog_gemini}
    }\hfil
    \subfloat[][Selected features depending on $\lambda$]{
        \includegraphics[width=0.45\linewidth]{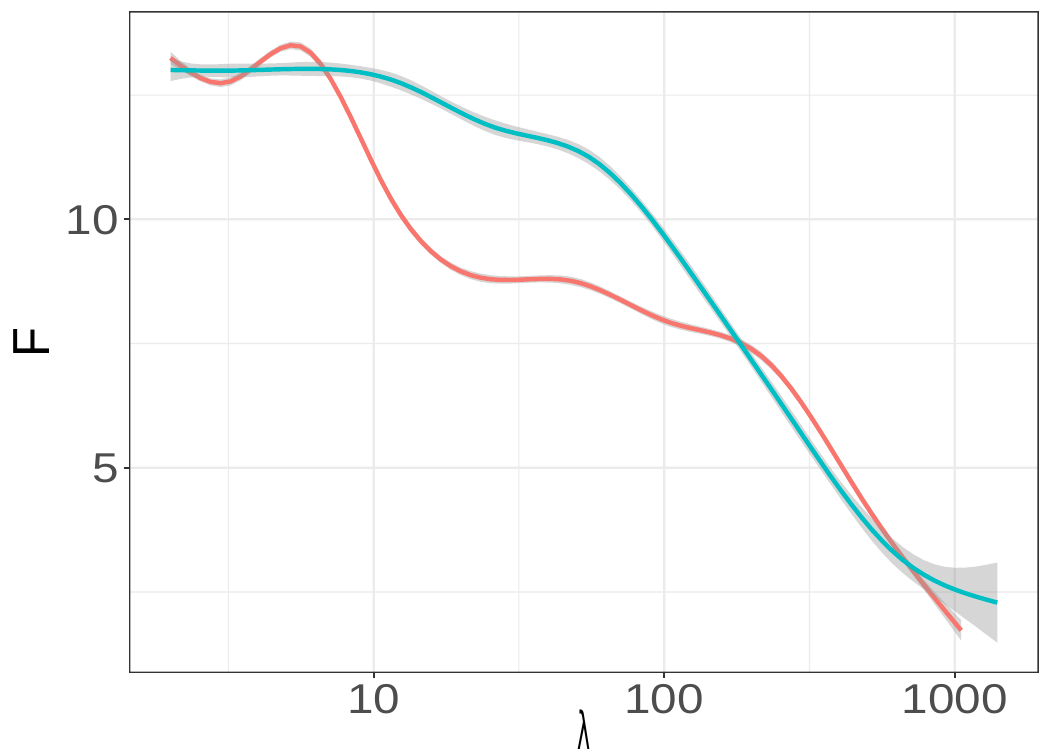}
        \label{sfig:heart_statlog_features}
    }
    \caption{Average training curves of Sparse GEMINI on the Heart Statlog dataset over 20 runs. Blue lines are Wasserstein, red lines are MMD.}
    \label{fig:heart_statlog}
\end{figure}

We ran Sparse GEMINI on two OpenML datasets that are often shown in related works: the US Congress dataset~\citep{almanac_98th_1984} and the Heart-statlog dataset~\citep{brown_diversity_2004}. The US Congress dataset describes the choice of the 435 representatives on 16 key votes in 1984. The labels used for evaluation are the political affiliations: 164 Republican against 267 Democrats. We replaced the missing values with 0 and converted the yes/no answers to 1, -1. Thus, an unknown label is equidistant from both answers. The Heart-statlog dataset describes 13 clinical and heart-related features with labels describing the presence or absence of cardiac disease among patients. We preprocessed it with standard scaling. For the US Congress dataset, we used one hidden layer of 20 nodes and a batch size of 87 samples. For the Heart-statlog dataset, we used 10 nodes and 90 samples. As we seek only two clusters, we only ran the one-vs-all versions of the GEMINI because it is strictly equal to the one-vs-one in binary clustering. Both datasets had a penalty increase of $\rho=10\%$. We first show the number of selected features evolving with $\lambda$ as well as the evolution of the GEMINI score as the number of features decreases respectively in Figure~\ref{fig:congress} for the US Congress dataset and in Figure~\ref{fig:heart_statlog} for Heart-statlog. Table~\ref{tab:openml_results} contains the performances for the two data sets, reporting the average number of variables selected over 20 runs according to our postprocessing selection criterion. We also added the performances of competitors from the previous section. However, we only managed to run Sparse Fisher EM on the Heart-statlog dataset, hence its absence for the US Congress scores. For comparison purposes, the best unsupervised accuracy reported on the Heart-statlog dataset by \citet{solorio-fernandez_review_2020} is 75.3\%, while Sparse GEMINI achieves 79\% with the MMD. The best score for all methods in the review~\citep{solorio-fernandez_review_2020} is 79.6\%, but this encompasses filter methods which Sparse GEMINI is not. We also get similar results to the best performances of~\citet{marbac_variable_2020} who report 33\% of ARI. Since most competitors retained all variables in the dataset, we chose to show as well the clustering performances without selection and hence with the greatest GEMINI score as well.

% \begin{table*}
%     \centering
%     \caption{ARI of Sparse GEMINI (OvA) on the Heart-statlog and US Congress datasets  with the average number of selected features. Standard deviation in subscript. Scores with an asterisk are the initial performances when using all features.}
%     \label{tab:openml_results}
%     \vskip 0.1in
%     \begin{tabular}{c c c c c c}
%     \toprule
%     &&\multicolumn{2}{c}{Heart-statlog}&\multicolumn{2}{c}{US Congress}\\
%     \cmidrule{3-6}
%     &&ARI&\# Variables&ARI&\# Variables\\
%     \midrule
%     \multicolumn{2}{c}{SparseKMeans}&0.18 \std{0.00}&13 \std{0.00}&\textbf{ 0.54 \std{0.00}}&16 \std{0.0}\\
%     \multicolumn{2}{c}{Clustvarsel}&0.03 \std{0.00}&2 \std{0.00}&0.00 \std{0.00}&2 \std{0.00}\\
%     \multicolumn{2}{c}{VSCC}&0.27 \std{0.00}&13 \std{0.00}&0.40 \std{0.00}&11 \std{0.00}\\
%     \multicolumn{2}{c}{Sparse Fisher EM}&0.19 \std{0.00}&1 \std{0.00}&-&-\\
%     \midrule
%     \multirow{2}{*}{Logistic regression}&MMD&\textbf{ 0.37 \std{0.03}}&7.5 \std{0.51}&\textbf{ 0.53 \std{0.02}}&8.3 \std{0.81}\\
%     &Wasserstein&0.33 \std{0.08}&5.8 \std{2.09}&0.48 \std{0.00}&8.0 \std{0.92}\\
%     \cmidrule{2-6}
%     \multirow{2}{*}{MLP}&MMD&0.32 \std{0.01}&8.0 \std{0.00}&0.48 \std{0.00}&3.1 \std{0.37}\\
%     &Wasserstein&0.32 \std{0.09}&8.4 \std{2.70}&0.47 \std{0.00}&2.0 \std{0.00}\\
%     \midrule
%     \multirow{2}{*}{MLP}&MMD\textsuperscript{*}&0.37 \std{0.02}&13 -&0.55 \std{0.01}&16 -\\
%     &Wasserstein\textsuperscript{*}&0.33 \std{0.09}&13 -&0.55 \std{0.02}&16 -\\
%     \bottomrule
%     \end{tabular}
% \end{table*}

\begin{table*}
    \centering
    \caption{ARI of Sparse GEMINI (OvA) on the Heart-statlog and US Congress datasets  with the average number of selected features. Standard deviation in subscript. Scores with an asterisk are the initial performances when using all features.}
    \label{tab:openml_results}
    \vskip 0.1in
    \begin{tabular}{c c c c c c}
    \toprule
    &&\multicolumn{2}{c}{Heart-statlog}&\multicolumn{2}{c}{US Congress}\\
    \cmidrule{3-6}
    &&ARI&\# Variables&ARI&\# Variables\\
    \midrule
    \multicolumn{2}{c}{SparseKMeans}&0.18 \std{0.00}&13 \std{0.00}&\textbf{ 0.54 \std{0.00}}&16 \std{0.0}\\
    \multicolumn{2}{c}{Clustvarsel}&0.03 \std{0.00}&2 \std{0.00}&0.00 \std{0.00}&2 \std{0.00}\\
    \multicolumn{2}{c}{VSCC}&0.27 \std{0.00}&13 \std{0.00}&0.40 \std{0.00}&11 \std{0.00}\\
    \multicolumn{2}{c}{Sparse Fisher EM}&0.19 \std{0.00}&1 \std{0.00}&-&-\\
    \midrule
    Logistic&MMD&\textbf{ 0.37 \std{0.03}}&7.5 \std{0.51}&\textbf{ 0.53 \std{0.02}}&8.3 \std{0.81}\\
    regression&Wasserstein&0.33 \std{0.08}&5.8 \std{2.09}&0.48 \std{0.00}&8.0 \std{0.92}\\
    \cmidrule{2-6}
    \multirow{2}{*}{MLP}&MMD&0.32 \std{0.01}&8.0 \std{0.00}&0.48 \std{0.00}&3.1 \std{0.37}\\
    &Wasserstein&0.32 \std{0.09}&8.4 \std{2.70}&0.47 \std{0.00}&2.0 \std{0.00}\\
    \midrule
    \multirow{2}{*}{MLP}&MMD\textsuperscript{*}&0.37 \std{0.02}&13 -&0.55 \std{0.01}&16 -\\
    &Wasserstein\textsuperscript{*}&0.33 \std{0.09}&13 -&0.55 \std{0.02}&16 -\\
    \bottomrule
    \end{tabular}
\end{table*}

We averaged the number of times each feature was selected according to the model over the 20 runs and sorted them decreasingly. This post-process revealed that the Wasserstein objective consistently selected the El Salvador Aid and the Aid to Nicaraguan Contras votes as sufficient to perform clustering. Indeed, these two votes are among the most discriminating features between Republicans and Democrats and were often chosen by other model-based methods~\citep{fop_variable_2018}. The MMD objective only added the Physician fee freeze vote to this subset.
Regarding the Heart-Statlog dataset, the MMD consistently picked a subset of 8 features out of 13, including, for example, age or chest pain type as relevant variables. Contrarily, the Wasserstein objective did not consistently choose the same subset of variables, yet its top variables that were selected more than 80\% of the runs agree with the MMD selection as well.
%On average, the MMD and Wasserstein models selected for the Heart-statlog the slope of the peak exercise ST segment, the number of major vessels colored by fluoroscopy and thal~\textbf{TODO: more}. Age and sex were among the first features to be often discarded. For the US congress dataset, the MMD tended to perform its clustering on the votes for the synfuels corporation cutback and the education spending, whereas the Wasserstein based its clustering on the aid to the Nicaraguan cutback and the MX-missile votes. These two votes and the education spending are among the most discriminating features between Republicans and Democrats and were often chosen by other model-based methods~\citep{fop_variable_2018}.

\subsubsection{Scalability example with the Prostate-BCR dataset}
\label{sssec:bcr_experiments}

% \begin{table*}
%     \centering
%     \caption{ARI scores of the Prostate BCR dataset for various numbers of clusters depending on the chosen type of targets. We either use the expected targets (BCR) regarding cancer prediction, or data source targets that identify the data origin of each sample. The indicated GEMINIs are in the one-vs-all setting.}
%     \label{tab:prostate_bcr_results}
%     \vskip 0.1in
%     \begin{tabular}{c c c c c c}
%         \toprule
%         \multicolumn{3}{c}{Model} & \multirow{2}{*}{\#Var} &\multirow{2}{*}{BCR targets ARI}&\multirow{2}{*}{Data source targets ARI}\\
%         \cmidrule{1-3}
%         Architecture&GEMINI&$K$\\
%         \midrule
%         \multirow{4}{*}{\begin{minipage}[t]{5em}\centering Logistic\\regression\end{minipage}}&\multirow{2}{*}{MMD}&2&810 \std{590}&-0.01 \std{0.00}&0.79 \std{0.01}\\
%         &&3&1229 \std{2270}&0.04 \std{0.00}&\textbf{ 1.00 \std{0.01}}\\
%         \cmidrule{4-6}
%         &\multirow{2}{*}{Wasserstein}&2&1334 \std{2561}&0.01 \std{0.02}&0.60 \std{0.13}\\
%         &&3&1430 \std{3127}&0.04 \std{0.01}&0.96 \std{0.06}\\
%         \cmidrule{2-6}
%         \multirow{4}{*}{MLP}&\multirow{2}{*}{MMD}&2&4013 \std{6541}&-0.01 \std{0.00}&0.78 \std{0.01}\\
%         &&3&4287 \std{6590}&0.03 \std{0.02}&0.93 \std{0.11}\\
%         \cmidrule{4-6}
%         &\multirow{2}{*}{Wasserstein}&2&4403 \std{6843}&0.00 \std{0.00}&0.65 \std{0.04}\\
%         &&3&4331 \std{6742}&0.02 \std{0.02}&0.80 \std{0.20}\\
%         \bottomrule
%     \end{tabular}
% \end{table*}

\begin{table*}
    \centering
    \caption{ARI scores of the Prostate BCR dataset for various numbers of clusters depending on the chosen type of targets. We either use the expected targets (BCR) regarding cancer prediction, or data source targets that identify the data origin of each sample. The indicated GEMINIs are in the one-vs-all setting.}
    \label{tab:prostate_bcr_results}
    \vskip 0.1in
    \begin{tabular}{c c c c c c}
        \toprule
        \multicolumn{3}{c}{Model} & \multirow{2}{*}{\#Var} & \multicolumn{2}{c}{Targets ARI}\\
        \cmidrule{1-3}
        Architecture&GEMINI&$K$&&BCR&\\
        \midrule
        \multirow{4}{*}{\begin{minipage}[t]{5em}\centering Logistic\\regression\end{minipage}}&\multirow{2}{*}{MMD}&2&810 \std{590}&-0.01 \std{0.00}&0.79 \std{0.01}\\
        &&3&1229 \std{2270}&0.04 \std{0.00}&\textbf{ 1.00 \std{0.01}}\\
        \cmidrule{4-6}
        &\multirow{2}{*}{Wasserstein}&2&1334 \std{2561}&0.01 \std{0.02}&0.60 \std{0.13}\\
        &&3&1430 \std{3127}&0.04 \std{0.01}&0.96 \std{0.06}\\
        \cmidrule{2-6}
        \multirow{4}{*}{MLP}&\multirow{2}{*}{MMD}&2&4013 \std{6541}&-0.01 \std{0.00}&0.78 \std{0.01}\\
        &&3&4287 \std{6590}&0.03 \std{0.02}&0.93 \std{0.11}\\
        \cmidrule{4-6}
        &\multirow{2}{*}{Wasserstein}&2&4403 \std{6843}&0.00 \std{0.00}&0.65 \std{0.04}\\
        &&3&4331 \std{6742}&0.02 \std{0.02}&0.80 \std{0.20}\\
        \bottomrule
    \end{tabular}
\end{table*}

To show the scalability of Sparse GEMINI, we also demonstrate its performance on the Prostate-BCR dataset, taken from \citet{vittrant_identification_2020}\footnote{The dataset is publicly available at \url{https://github.com/ArnaudDroitLab/prostate_BCR_prediction}}. This dataset is a combination of transcriptomics data from 3 different sources: the Cancer Genom Atlas~\citep{abeshouse_molecular_2015}, the GSE54460 dataset from the NCBI website, and the PRJEB6530 project of the European Nucleotide Archive. The combined dataset contains 25,904 transcripts over 171 filtered patients with long-term follow-up, counting 52, 96 and 23 patients from the respective sources. The objective is to find biochemical recurrences (BCR) of prostate cancer through the transcriptomic signature, hence binary targets.

To carefully eliminate the variables, we increase $\lambda$ gradually by 2\%. We took a simple MLP with only one hidden layer of 100 neurons. We chose to run until we reached 400 features or less, following \citet{vittrant_identification_2020}. We trained Sparse GEMINI with OvA objectives 5 times to find either 2 or 3 clusters in order to break down possible substructures among the supervised targets.  %For the evaluation of the 3 clusters case, we binarised the results by mapping each cluster to the class in which it had the most samples 

\begin{table*}
    \centering
    \caption{ARI scores of the Prostate BCR dataset for various numbers of clusters depending on the chosen type of targets. We either use the expected targets (BCR) regarding cancer prediction, or data source targets that identify the data origin of each sample. The indicated GEMINIs are in the one-vs-all setting.}
    \label{tab:prostate_bcr_results_full_features}
    \vskip 0.1in
    \begin{tabular}{c c c c c}
        \toprule
        \multicolumn{3}{c}{Model} & \multirow{2}{*}{BCR targets ARI}&\multirow{2}{*}{Data source targets ARI}\\
        \cmidrule{1-3}
        Architecture&GEMINI&$K$\\
        \midrule
        \multirow{4}{*}{\begin{minipage}[t]{5em}\centering Logistic\\regression\end{minipage}}&\multirow{2}{*}{MMD}&2&-0.01 \std{0.00} & 0.80 \std{0.01}\\
        &&3& 0.03\std{0.01} & 0.97 \std{0.07}\\ 
        \cmidrule{4-5}
        &\multirow{2}{*}{Wasserstein}&2&0.00 \std{0.00} & 0.75 \std{0.00}\\
        &&3& 0.03 \std{0.02} & 0.90 \std{0.13}\\
        \cmidrule{2-5}
        \multirow{4}{*}{MLP}&\multirow{2}{*}{MMD}&2& 0.01 \std{0.02} & 0.56 \std{0.11}\\
        &&3& 0.04 \std{0.01} & 0.95 \std{0.05}\\
        \cmidrule{4-5}
        &\multirow{2}{*}{Wasserstein}&2 & 0.00 \std{0.00} & 0.68 \std{0.07} \\
        &&3& 0.03 \std{0.03} & 0.86 \std{0.16}\\
        \bottomrule
    \end{tabular}
\end{table*}

Interestingly, we observed in Table~\ref{tab:prostate_bcr_results} that the clustering results did not catch up with the actual BCR targets, with an ARI close to 0 most of the time. However, upon evaluation of the clusters with respect to the original source of each sample, we found scores close to 1 of ARI in the case of the MMD GEMINI. Thus, the unsupervised algorithm was able to find sufficient differences in distribution between each data source to discriminate them. Additionally, consistent subsets of features were always selected as the final subset on all 5 runs depending on the GEMINI. This implies that even without the best GEMINI within a range for feature selection, several runs can lead to identifying subsets of relevant data as well. This example illustrates how even in the presence of potentially legitimate labels, there exist other valid cluster structures in the data~\citep{hennig_what_2015}. By comparing these results with the performances using all features in Table~\ref{tab:prostate_bcr_results_full_features}, we observe that in general the MMD objective gained in ARI with the decrease of the number of features, whereas Wasserstein GEMINI is stronger when observing all features.

These results can be viewed as discovering batch effect in the data. Batch effect, also known as batch variation, is a phenomenon that occurs in biological experiments where the results are affected by factors unrelated to the experimental variables being studied. These factors can include variations in sample processing, measurement conditions, people manipulating the samples, or equipment used. One common example of a batch effect is observed in microarray or RNA sequencing experiments, where the samples are processed in different batches and the results are affected by variations in the reagents or protocols used. Batch effects in microarray experiments have been shown to originate from multiple causes, including variations in the labelling and hybridisation protocols used, leading to differences in the intensity of gene expression signals~\citep{luo_comparison_2010}.

To minimise batch effects, it is important to control variables such as reagents, protocols, and equipment used, and to use appropriate normalisation and data analysis methods to account for these variations. Several approaches can be used to detect batch effects in RNA-seq experiments, including PCA ~\citep{reese_new_2013} and clustering. For this latter, Hierarchical clustering is often used as a method that groups samples based on their similarity in gene expression patterns, and batch effects can be identified based on dendrogram analysis~\citep{leek_tackling_2010}.

\subsection{Discussion}
\label{sec:discussion}

Our first observation from Table~\ref{tab:synthetic_datasets_results} is that the Sparse GEMINI algorithm can achieve performance close to some competitors in terms of ARI while performing better in variable selection, especially for the one-vs-one MMD. The MMD is a distance computed between expectations, making it thus insensitive to small variations of the kernel, typically when noisy variables are introduced, contrary to the Wasserstein distance which takes a global point of view on the distribution. Specifically, the algorithm is good at discarding noisy variables, but less competitive with regard to redundant variables as illustrated with the second synthetic dataset. Nonetheless, the ARI remains competitive even though the model failed to give the correct ground for the clustering.

Additionally, the training path produces critical values of $\lambda$ at which the features disappear. Thus, the algorithm produces an explicit unsupervised metric of the relevance of each feature according to the clustering. Typically, plateaus of the number of used variables like in figures~\ref{sfig:congress_features} and~\ref{sfig:heart_statlog_features} for the MMD shed light on different discriminating subsets.  We also find that the empirical threshold of 90\% of the maximal GEMINI to select fewer variables is an efficient criterion. In case of a too sudden collapse of variables, we encourage training over again models on iteratively selected subsets of features. Indeed, as $\lambda$ increases during training, the collapse of the number of selected variables will often happen when the geometric increase is too strong which might lead to unstable selections.

%%%%%%%%%%%%%%%%%%%%%%%%%%%%%%%%%%%%%%%%%%%%%%%%%%%%%%%%%%%%%%%%%%%%%%%%%
% CONCLUSION
%%%%%%%%%%%%%%%%%%%%%%%%%%%%%%%%%%%%%%%%%%%%%%%%%%%%%%%%%%%%%%%%%%%%%%%%%

\section{Conclusion}
\label{sec:conclusion}

We presented a novel algorithm named Sparse GEMINI that jointly performs clustering and feature selection by combining GEMINI for objective and an $\ell_1$ penalised skip connection. The algorithm shows good performances in eliminating noisy irrelevant variables while maintaining relevant clustering. To eliminate redundant variables, our future works can focus on adding the correlation of selected variables in the penalty. Owing to the nature of multi-layered perceptrons, Sparse GEMINI is easily scalable to high-dimensional data and thus provides an unsupervised technique to get a projection of the data. However, the limits of the scalability are the number of clusters and samples per batch due to the complex nature of GEMINI. Thus, we believe that Sparse GEMINI is a relevant algorithm for multi-omics data where the number of samples is often little and the number of features large, especially when it is hard to design a good generative model for such data. As a concluding remark, we want to draw again the attention to the discriminative nature of the algorithm: Sparse GEMINI focuses on the design of a decision boundary instead of parametric assumptions.

%%%%%%%%%%%%%%%%%%%%%%%%%%%%%%%%%%%%%%%%%%%%%%%%%%%%%%%%%%%%%%%%%%%%%%%%%
% BACKMATTER
%%%%%%%%%%%%%%%%%%%%%%%%%%%%%%%%%%%%%%%%%%%%%%%%%%%%%%%%%%%%%%%%%%%%%%%%%
\backmatter

\bmhead{Acknowledgements}
This work has been supported by the French government, through the 3IA C\^ote d'Azur, Investment in the Future, project managed by the National Research Agency (ANR) with the reference number ANR-19-P3IA-0002. We would also like to thank the France Canada Research Fund (FFCR) for their contribution to the project. This work was partly supported by EU Horizon 2020 project AI4Media, under contract no. 951911. Finally, it was also supported by the Health Data Hub.

\section*{Declarations} % See requirements in https://www.springer.com/journal/11222/submission-guidelines

\begin{itemize}
    \item Author L.Ohl received funding from the French government, through the 3IA C\^ote d'Azur, Investment in the Future, project managed by the National Research Agency (ANR) with the reference number ANR-19-P3IA-0002.
    \item This work benefited as well from the EU Horizon 2020 project AI4Media, under contract no. 951911.
    \item This project was as well funded by the France Canada Research Fund (FFCR) to all authors of the paper except C. Bouveyron.
\end{itemize}

We do not have any employment or financial interest emerging from this paper.

%%%%%%%%%%%%%%%%%%%%%%%%%%%%%%%%%%%%%%%%%%%%%%%%%%%%%%%%%%%%%%%%%%%%%%%%%
% APPENDICES
%%%%%%%%%%%%%%%%%%%%%%%%%%%%%%%%%%%%%%%%%%%%%%%%%%%%%%%%%%%%%%%%%%%%%%%%%

\begin{appendices}

\section{The dynamic training regime}
\label{app:dynamic_training}
As features get eliminated during the training, the notion of affinity (distance $\delta$ or kernel $\kappa$) and clustering with respect to GEMINI between two samples changes. Indeed, GEMINI aims at maximising a distance between two related distributions using an affinity computed between samples, yet removing features from the inference implies that we no longer cluster the same original data space, but rather a subspace at step $t$: $\mathcal{X}_t = \prod_{j \in I_t} \mathcal{X}_j$. If we still compute our affinity function using all features from $\mathcal{X}$ the extra removed features may bring noise compared to the affinity between the relevant features, and thus bring confusion with regards to the ideal decision boundary.

To respect the original notion of GEMINI in clustering, we introduce the dynamic training regime, where at each time step $t$, the affinity function is computed using only the subset of relevant features $I_t$. We call \emph{static} regime the training with usage of all features in the affinity function as described in section~\ref{sec:method}. The advantage of the dynamic training regime is that it respects the notion of GEMINI with regard to the decision boundary, while the static regime yields comparable values of GEMINI independently of the number of selected features. However, the dynamic regime is incompatible with the selection process described in section~\ref{ssec:selection_process} because any change of data space implies a change of values for kernels or distances and thus for GEMINI, making models incomparable. Moreover, we may have more theoretical guarantees of convergence for the usual static regime than in the dynamic regime which may seem unstable.

We experiment this approach again with the synthetic data sets and report the results in Table~\ref{tab:synthetic_dynamic}. For this experiment, we only evaluated the performances on the final subset of selected features. However, since Sparse GEMINI is trained until a user-defined number of features is reached, we avoid unfair comparisons with other variable selection methods and do not report the VSER and the CVR. Our main observation on the introduction of the dynamic regime is that it greatly improves the clustering performances of the Wasserstein-GEMINI while not affecting the MMD-GEMINI. This success can be explained by the removal of variables as the removal of noise in the distance computation which is crucial for the Wasserstein distance because it takes a global point of view on the complete distribution. In contrast, the MMD only considers the expectation, which helps in removing noisy variations of the distance around informative variables.

\begin{table*}
    \centering
    \caption{ARI scores on the synthetic datasets with the dynamic regime of training for the Sparse GEMINI using MLPs}
    \label{tab:synthetic_dynamic}
    \begin{tabular}{c c c c c}
        \toprule
        \multirow{2}{*}{Method}&\multicolumn{2}{c}{MMD}&\multicolumn{2}{c}{Wasserstein}\\
        \cmidrule{2-5}
        &OvA&OvO&OvA&OvO\\
        \midrule
        Scenario 1&0.12 (0.13)&\textbf{ 0.15 (0.12)}&0.05 (0.09)&0.07 (0.06)\\
        Scenario 2&0.48 (0.11)&\textbf{ 0.63 (0.21)}&0.37 (0.11)&0.30 (0.13)\\
        Scenario 3&\textbf{ 0.23 (0.04)}&0.20 (0.03)&0.11 (0.05)&0.11 (0.06)\\
        Scenario 4&0.45 (0.07)&\textbf{ 0.88 (0.03)}&0.82 (0.13)&0.85 (0.10)\\
        Scenario 5&0.62 (0.09)&\textbf{ 0.84 (0.05)}&0.56 (0.20)&0.48 (0.17)\\
        \midrule
        Dataset 2&0.54 (0.04)&0.54 (0.05)&0.50 (0.08)&\textbf{ 0.56 (0.01)}\\
        \bottomrule
    \end{tabular}
\end{table*}

\section{Differentiation of the OvA MMD}
\label{app:mmd_ova_computations}

\subsection{Alternative computation of the forward pass}

We consider the computations starting from a row-stochastic matrix $\vec{\tau}\in\mathbb{R}^{N\times K}$, typically the softmax output of a model. We focus here only on the computations of the objective function, the OvA MMD. First, we can compute the cluster proportions:

\begin{equation}
\vec{\pi} = \frac{1}{N}\vec{1}_N^\top\vec{\tau}.
\end{equation}

Our goal is to compute the vector $\vec{\Delta}\in\mathbb{R}^{K}$ where the $k$-the component is the squared distance in the Hilbert space between one cluster distribution and the data distribution:

\begin{equation}
\vec{\Delta}_k = \sum_{i,j}^{N,N} \tilde{\kappa}_{i,j} \left[\frac{\vec{\tau}_{ki}\vec{\tau}_{kj}}{\pi_k^2} + 1 -2 \frac{\vec{\tau}_{ki}}{\pi_k}\right].
\end{equation}

To that end, we introduce the matrix $\vec{\alpha}\in\mathbb{R}^{N\times K}$ which is the element-wise division of $\vec{\tau}$ by the proportions of the matching cluster.

\begin{equation}
\vec{\alpha} = \vec{\tau}\oslash (\vec{1}_N\vec{\pi}^\top) = \left[\frac{\vec{\tau}_{ki}}{\pi_k}\right].
\end{equation}

Individually, we can interprete the value of $\alpha_ik$ as the ratio $p(y=k|\vec{x}_i)/p(y=k)$ or $p(\vec{x}_i|y=k)/p(\vec{x}_i)$. This represents the relative strength of the sample in the cluster distribution. We can then deduce the writing of $\vec{\Delta}$:

% \begin{align}
% \vec{\Delta} &= \text{diag}\left(\vec{\alpha}^\top\vec{\tilde{\kappa}}\vec{\alpha}\right) + \vec{1}_{K\times N}\vec{\tilde{\kappa}}\vec{1}_N - 2 \vec{\alpha}^\top\vec{\tilde{\kappa}}\vec{1}_{N},\\
% &=\vec{a}+\vec{c}-2\vec{b}.
% \end{align}

% \begin{align}
% \vec{\Delta} &= \text{diag}\left(\vec{\alpha}^\top\vec{\tilde{\kappa}}\vec{\alpha}\right) + \vec{1}_{K\times N}\vec{\tilde{\kappa}}\vec{1}_N - 2 \vec{\alpha}^\top\vec{\tilde{\kappa}}\vec{1}_{N},\\
% &=\vec{a}+\vec{c}-2\vec{b}.
% \end{align}

To name the elements, $\vec{a}$ is the cluster contribution: how the distribution of the cluster contributes to increase the distance, and the same goes for the constant $\vec{c}$ which represents the agnostic data strength. Finally, $\vec{b}$ is the agreement between the two distributions $p(y|=k|\vec{x})$ and $p(\vec{x})$ which diminishes the value of the MMD: the more the cluster distribution takes to the data (by having everything in the same cluster), the stronger $\vec{b}$ is and the lower the MMD. Yet, to simplify the derivatives to compute later, we introduce two intermediary variables:

\begin{equation}
\vec{\gamma} = \vec{\tilde{\kappa}}\vec{\alpha},
\end{equation}
and

\begin{equation}
\vec{\omega} = \vec{\alpha}^\top\vec{\gamma},
\end{equation}
of respective shapes $N\times K$ and $K\times K$. Thus, we simply rewrite:

\begin{align}
\vec{\Delta} &= \vec{a}+\vec{c}-2\vec{b}, \\&= \text{diag}(\vec{\omega}) +  \vec{1}_{K\times N}\vec{\tilde{\kappa}}\vec{1}_N  - 2 \vec{\gamma}^\top\vec{1}_N.
\end{align}

Finally, assuming the square root is applied element-wise, we can write the final objective as:

\begin{equation}
\hat{\I}^\text{ova}_\text{MMD}(\vec{x},y|\theta) = \vec{\pi}^\top \sqrt{\vec{\Delta}}.
\end{equation}

\begin{figure}
\centering
\begin{tikzpicture}[every node/.style = {circle, thick, draw}]
\node (tau) {$\tau$};
\node[below right= of tau] (pi) {$\pi$};
\node[right= of pi] (alpha) {$\alpha$};
\node[right= of alpha] (kappa) {$\kappa$};
\node[below right = of alpha] (gamma) {$\gamma$};
\node[below right= of alpha, below left= of kappa] (omega) {$\omega$};
\node[below= of gamma] (b) {$b$};
\node[below= of omega] (a) {$a$};
\node[right= of b] (c) {$c$};
\node[below= of a] (delta) {$\Delta$};
\node[left= of delta] (i) {$\I$};

\draw[thick,->] (tau) -- (pi);
\draw[thick,->] (tau) -- (alpha);
\draw[thick,->] (pi) -- (alpha);
\draw[thick,->] (alpha) -- (gamma);
\draw[thick,->] (alpha) -- (omega);
\draw[thick,->] (gamma) -- (b);
\draw[thick,->] (gamma) -- (omega);
\draw[thick,->] (kappa) -- (gamma);
\draw[thick,->] (kappa) -- (c);
\draw[thick,->] (omega) -- (a);
\draw[thick,->] (a) -- (delta);
\draw[thick,->] (b) -- (delta);
\draw[thick,->] (c) -- (delta);
\draw[thick,->] (delta) -- (i);
\draw[thick,->] (pi) -- (i);
\end{tikzpicture}
\caption{Summary of computations for the forward pass of the OvA MMD using matrices}
\label{fig:forward_pass}
\end{figure}
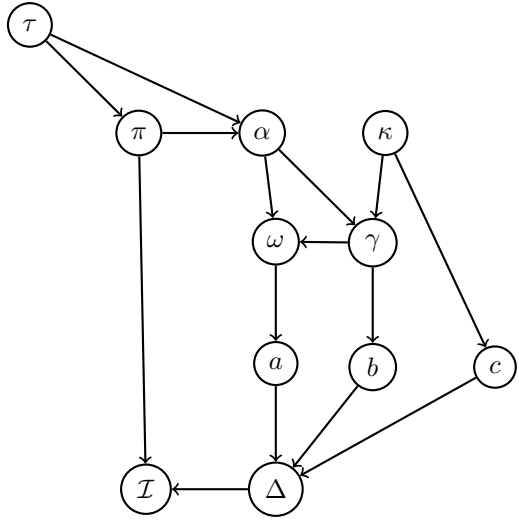

The graph of computations is summarised in Figure~\ref{fig:forward_pass}.

\subsection{Backward pass}

We can now compute the derivatives of each part of the graph with respect to the conditional probabilities: $\diff{\hat{\I}}{\vec{\tau}}$. By reversing the graph, we get the list of the following sorted derivatives to compute: (1) $\diff{\hat{\I}}{\vec{\Delta}}$
(2) $\diff{\hat{\I}}{\vec{a}}$
(3) $\diff{\hat{\I}}{\vec{b}}$
(4) $\diff{\hat{\I}}{\vec{\omega}}$
(5) $\diff{\hat{\I}}{\vec{\gamma}}$
(6) $\diff{\hat{\I}}{\vec{\alpha}}$
(7) $\diff{\hat{\I}}{\vec{\pi}}$
(8) $\diff{\hat{\I}}{\vec{\tau}}$.

To compute these derivatives, we will follow an automatic differentiation procedure. All derivatives correspond to the gradient of a scalar with respect to a matrix or vector, hence all derivatives will keep the same shape as the denominator. Notice that because $c$ and $\tilde{\kappa}$ do not depend on $\vec{\tau}$, they will not produce any gradient.

\subsubsection{Deriving for \texorpdfstring{$\Delta$}{Delta}}

We simply take the vector $\pi$ for this derivative that summed the square root of $\Delta$. Additionally, the element-wise square root is differentiated. Thus:

\begin{equation}
\diff{\hat{\I}}{\vec{\Delta}} = \frac{\vec{\pi}}{2\sqrt{\Delta}}.
\end{equation}
\subsubsection{Deriving for \texorpdfstring{$\vec{a}$}{a} and \texorpdfstring{$\vec{b}$}{b}}
The contributions of $\vec{a}$ and $\vec{b}$ are simple element-wise sums of vectors. They have the same shape as the previous gradient. Therefore:

\begin{equation}
\diff{\hat{\I}}{\vec{a}} = \diff{\hat{\I}}{\vec{\Delta}} = \frac{\vec{\pi}}{2\sqrt{\Delta}},
\end{equation}

\begin{equation}
\diff{\hat{\I}}{\vec{b}} = -2\diff{\hat{\I}}{\vec{\Delta}} = -\frac{\vec{\pi}}{\sqrt{\Delta}}.
\end{equation}

\subsubsection{Deriving for \texorpdfstring{$\omega$}{omega}}

Since we took only the diagonal of $\omega$ for computation, the gradient with respect to $\omega$ will be a diagonal matrix, which diagonal is exactly the previously committed error:

\begin{equation}
\diff{\hat{\I}}{\vec{\omega}} = \text{diag}\left(\diff{\hat{\I}}{\vec{a}}\right) = \frac{1}{2}\text{diag}\left(\frac{\vec{\pi}}{\sqrt{\vec{\Delta}}}\right).
\end{equation}

\subsubsection{Deriving for \texorpdfstring{$\gamma$}{gamma}}

The vector $\gamma$ contributed two times in the computation graph: once to $\vec{\omega}$ and another time for $\vec{b}$. Both cases involve simple matrix multiplications. We can sum the matrix gradient to both errors to get:

\begin{align}
\diff{\hat{\I}}{\vec{\gamma}} &= \diff{\hat{\I}}{\vec{\omega}}\vec{\alpha}^\top + \diff{\hat{\I}}{\vec{b}}\vec{1}_N^\top,\\
&=\frac{1}{2}\vec{\alpha}\text{diag}\left(\frac{\vec{\pi}}{\sqrt{\vec{\Delta}}}\right) - \vec{1}_N\left(\frac{\vec{\pi}}{\sqrt{\vec{\Delta}}}\right)^\top.
\end{align}

Notice that for the derivative from $\vec{b}$, we had to transpose the error since $\vec{b}$ is computed using $\vec{\gamma}^\top$. Here, we can remark that by unfolding the definition of $\vec{\alpha}$, and thanks to matrix product with the diagonal product, the values of $\vec{\pi}$ gets cancelled. Therefore:

\begin{equation}
\diff{\hat{\I}}{\vec{\gamma}} = \frac{1}{2}\frac{\vec{\tau}}{\vec{1}_N\sqrt{\vec{\Delta}}^\top}- \vec{1}_N\left(\frac{\vec{\pi}}{\sqrt{\vec{\Delta}}}\right)^\top.
\end{equation}

\subsubsection{Deriving for \texorpdfstring{$\alpha$}{alpha}}

As we did for $\vec{\gamma}$, we need to sum here the gradient contributions of $\vec{\alpha}$ to $\vec{\omega}$ and $\vec{\gamma}$. Both are matrix multiplications, however we add a transposition in the case of $\omega$. Thus:

\begin{align}
\diff{\hat{\I}}{\vec{\alpha}} &= \left(\diff{\hat{\I}}{\vec{\omega}}\vec{\gamma}^\top\right)^\top + \vec{\tilde{\kappa}}^\top\diff{\hat{\I}}{\vec{\gamma}},\\
&=\frac{1}{2}\vec{\gamma}\text{diag}\left(\frac{\vec{\pi}}{\sqrt{\vec{\Delta}}}\right) + \vec{\tilde{\kappa}} \left(\frac{1}{2}\frac{\vec{\tau}}{\vec{1}_N\sqrt{\vec{\Delta}}^\top}- \vec{1}_N\left(\frac{\vec{\pi}}{\sqrt{\vec{\Delta}}}\right)^\top\right).
\end{align}

Here, we can unfold the definition of $\vec{\gamma}$ to make a common factor $\vec{\tilde{\kappa}}$ appear on the left matrix multiplication. Thus:

\begin{equation}
\diff{\hat{\I}}{\vec{\alpha}} = \vec{\tilde{\kappa}}\left[\frac{1}{2}\vec{\alpha}\text{diag}\left(\frac{\vec{\pi}}{\sqrt{\vec{\Delta}}}\right) +\frac{1}{2}\frac{\vec{\tau}}{\vec{1}_N\sqrt{\vec{\Delta}}^\top} - \vec{1}_N\left(\frac{\vec{\pi}}{\sqrt{\vec{\Delta}}}\right)^\top\right].
\end{equation}

We notice the exact same simplification on the left term between $\vec{\alpha}$ and the diagonal matrix as we had for the gradient w.r.t. $\vec{\gamma}$. Rewriting this term is exactly equal to the second, and thus:

\begin{equation}
\diff{\hat{\I}}{\vec{\alpha}} = \vec{\tilde{\kappa}}\left[\frac{\vec{\tau}}{\vec{1}_N\sqrt{\vec{\Delta}}^\top}-\vec{1}_N\left(\frac{\vec{\pi}}{\sqrt{\vec{\Delta}}}\right)^\top\right].
\end{equation}

\subsubsection{Deriving for \texorpdfstring{$\pi$}{pi}}

Same procedure for $\vec{\pi}$ by summing the contributions of the gradient from $\vec{\alpha}$ and the dot product with $\sqrt{\vec{\Delta}}$ in $\I$. For the derivative from $\vec{\alpha}$, we multiply the rows of the previous error by the squared inverse of $\vec{\pi}$ and $\vec{\beta}$ and sum them. Hence:

\begin{align}
\diff{\hat{\I}}{\vec{\pi}} &= \sqrt{\vec{\Delta}} - \left[\left[\frac{\vec{\tau}^\top}{\sqrt{\vec{\Delta}}\vec{1}_N^\top} - \frac{\vec{\pi}}{\sqrt{\vec{\Delta}}}\vec{1}_N^\top\right]\vec{\tilde{\kappa}}\odot \frac{\vec{\beta}}{\vec{\pi}^2\vec{1}_N^\top}\right]\vec{1}_N,\\
&= \sqrt{\vec{\Delta}} - \left[\left[\frac{\vec{\tau}^\top}{\sqrt{\vec{\Delta}}\vec{1}_N^\top} - \frac{\vec{\pi}}{\sqrt{\vec{\Delta}}}\vec{1}_N^\top\right]\vec{\tilde{\kappa}}\odot \frac{\vec{\alpha}}{\vec{\pi}\vec{1}_N^\top}\right]\vec{1}_N.
\end{align}

Since the inverse factor $1/\vec{\pi}\vec{1}_N^\top$ is constant row-wise, we can incorporate it directly to the left term of the matrix multiplication. This simplifies again the notations:

\begin{align}
\diff{\hat{\I}}{\vec{\pi}} &= \sqrt{\vec{\Delta}} + \left[\left( \frac{\vec{\alpha}^\top\vec{\tilde{\kappa}}}{\sqrt{\vec{\Delta}}\vec{1}_N^\top} - \frac{\vec{1}_N^\top\vec{\tilde{\kappa}}}{\sqrt{\vec{\Delta}}\vec{1}_N^\top}\right)\odot\vec{\alpha}\right]\vec{1}_N.
\end{align}

Here, the combination of the element-wise multiplication by $\vec{\alpha}$ followed by a sum over all samples is in fact equal to the respective distance terms $\vec{a}$ and $\vec{b}$. First first simplification yields:

\begin{align}
\diff{\hat{\I}}{\vec{\pi}} &= \sqrt{\vec{\Delta}} - \frac{\vec{a}-\vec{b}}{\sqrt{\Delta}}.
\end{align}

To finally go further, we can use the definition of $\vec{\Delta}$ to replace the left term by another. Indeed, since:

\begin{equation}
\vec{a}-\vec{b} = \vec{\Delta}+\vec{b}-\vec{c},
\end{equation}
and the denominator $\sqrt{\Delta}$ gets cancelled by $\vec{\Delta}$, we obtain:

\begin{equation}
\diff{\hat{\I}}{\vec{\pi}} = \frac{\vec{c}-\vec{b}}{\sqrt{\Delta}}.
\end{equation}

\subsubsection{Deriving for \texorpdfstring{$\tau$}{tau}}

Finally, the gradient for $\vec{\tau}$ sums contributions from both $\vec{\alpha}$ and $\vec{\pi}$.  In both cases, we just consider element-wise operations, so the global gradient will just be element-wise multiplication of the errors, with a specific repetition over all rows for the gradient from $\vec{\pi}$:

\begin{align}
\diff{\hat{\I}}{\vec{\tau}} &= \frac{\vec{1}_N}{N}\diff{\hat{\I}}{\vec{\pi}}^\top + \diff{\hat{\I}}{\vec{\alpha}}\odot \frac{1}{\vec{1}_N\vec{\pi}^\top},\\
&= \vec{1}_N {\frac{\vec{c}-\vec{b}}{N\sqrt{\Delta}}}^\top + \vec{\tilde{\kappa}}\left[\frac{\vec{\tau}}{\vec{1}_N\sqrt{\vec{\Delta}}^\top}-\vec{1}_N\left(\frac{\vec{\pi}}{\sqrt{\vec{\Delta}}}\right)^\top\right] \odot \frac{1}{\vec{1}_N\vec{\pi}^\top},\\
&= \vec{1}_N {\frac{\vec{c}-\vec{b}}{N\sqrt{\Delta}}}^\top + \vec{\tilde{\kappa}}\left[\frac{\vec{\alpha}}{\vec{1}_N\sqrt{\vec{\Delta}}^\top} - \frac{1}{\vec{1}_N\sqrt{\vec{\Delta}}^\top} \right].
\end{align}

To conclude, we can factorise all terms by the common denominator:

\begin{equation}
\diff{\hat{\I}}{\vec{\tau}} = \frac{1}{\vec{1}_N\sqrt{\vec{\Delta}}^\top} \odot \left[ \frac{\vec{1}_N}{N}(\vec{c}-\vec{b})^\top + \vec{\tilde{\kappa}} (\vec{\alpha}-\vec{1}_{N\times K}) \right].
\end{equation}

For further simplification of the gradients, we can unfold again the definition of $\vec{b}$ and $\vec{c}$ as follows:

\begin{align}
\frac{\vec{1}_N}{N}(\vec{c}-\vec{b})^\top &= \frac{1}{N}\left[\vec{1}_{N\times N}\vec{\tilde{\kappa}}\vec{1}_{N\times K} - \vec{1}_{N\times N}\vec{\tilde{\kappa}}\vec{\alpha}\right], \\
&= \frac{1}{N}\left[\vec{1}_{N\times N}\vec{\tilde{\kappa}}\left(\vec{1}_{N\times K}-\vec{\alpha}\right)\right].
\end{align}

Thus, we can conclude that the final equation for the gradient of the OvA MMD is:

\begin{equation}
\diff{\hat{\I}}{\vec{\tau}} = \frac{1}{\vec{1}_N\sqrt{\vec{\Delta}}^\top} \odot \left[ \left(I_{N}-\vec{1}_{N\times N}/N\right)\vec{\tilde{\kappa}}\left(\vec{\alpha}-\vec{1}_{N\times K}\right) \right].
\end{equation}

To be more precise, we can even express the value for a component at position $i,k$:

\begin{align}
\diff{\hat{\I}}{\vec{\tau}}_{i,k}&= \left[\frac{1}{\sqrt{\Delta}_k}\sum_{j=1}^N \left(\vec{\tilde{\kappa}}_{ij}-\frac{1}{N}\sum_{l=1}^N\vec{\tilde{\kappa}}_{jl}\right)(\alpha_{jk}-1)\right]\\
&= \left[\frac{1}{\sqrt{\Delta}_k} (c - b_k + \gamma_{ik} - \vec{\bar{\kappa}}_i)\right],
\end{align}
with $\vec{\bar{\kappa}}_i = \sum_{j=1}^N \tilde{\kappa}_{ij}$.

%TODO: conclusion

\section{Differentiation of the OvO MMD}
\label{app:mmd_ovo_computations}

\subsection{Forward pass}

We will proceed here to the exact same reasoning as in the OvA MMD. We first compute the distance $\vec{\Delta}$ before summing them with $\vec{\pi}$. Contrary to the OvA MMD, $\vec{\Delta}$ is now a matrix of shape $K\times K$ where each entry describes the distance between two clusters $k$ and $k^\prime$:

\begin{equation}
\hat{\I}_\text{MMD}^\text{ovo}(\vec{x},y|\theta) = \vec{\pi}^\top \sqrt{\vec{\Delta}}\vec{\pi}.
\end{equation}

As previously done, we can express the squared distance as the sum of two self-contributions minus a cross-contribution. These contributions will be here matrices of shape $K\times K$. Yet, we can notice that in the OvO MMD, the matrix $\vec{\Delta}$ is symmetric. Simply put, the cross-contribution is symmetric, and the two self-contributions are the transposed of each other:

\begin{align}
\vec{\Delta} &= \vec{A}+\vec{C} - 2\vec{B},\\
&= \vec{A} + \vec{A}^\top - 2\vec{B}.
\end{align}

We can here realise that the matrix $\vec{A}$ is in fact a column-wise copy of the vector $\vec{a}$ from the previous computations with OvA MMD. Similarly, $\vec{B}$ is the entire matrix $\vec{\omega}$ while $\vec{A}$ only consists in its diagonal. Therefore:

\begin{align}
\vec{\Delta} &= \text{diag}(\vec{\omega})\vec{1}_K^\top + \vec{1}_K\text{diag}(\vec{\omega})^\top - 2\vec{\omega}.
\end{align}

The remaining of the definition of $\vec{\omega}$ strictly unfolds from the OvA MMD forward pass.

\subsection{Backward pass}

In the specific case of the OvO, we have square roots of values of $\vec{\Delta}$ which can be equal to 0, hence undifferentiable. This is in fact not a burden since in principle, these 0 only happen when we evaluate the MMD between a cluster and itself. Thus, we can discard easily the null components of $\vec{\Delta}$ during the final sum (expectation over $\pi$) and adopt locally the small convention that the derivative of $\I$ w.r.t. $\vec{\Delta}$ will be equal to 0 on the diagonal, despite the square root computation.

\subsubsection{Deriving for \texorpdfstring{$\vec{\Delta}$}{Delta}}

We start simple, the derivative is simply a square matrix where all components are the cartesian product of the vector $\vec{\pi}$:

\begin{equation}
\diff{\hat{\I}}{\vec{\Delta}} = \frac{\pi\pi^\top}{2\sqrt{\Delta}}.
\end{equation}

From now on, we will arbitrarily say that $\left(\diff{\hat{\I}}{\vec{\Delta}}\right)_{k,k} = 0$ because it was not summed at the end of the forward pass in the OvO MMD. Thus, we will write for clarity:

\begin{equation}
\diff{\hat{\I}}{\vec{\Delta}} = \frac{\pi\pi^\top}{2\sqrt{\Delta}} \odot (\vec{1}_{K\times K}-\vec{I}_K).
\end{equation}

\subsubsection{Deriving for \texorpdfstring{$\vec{\omega}$}{omega}}

For the gradient w.r.t. $\vec{\omega}$, we have two contributions to sum, one which comes from the diagonal of $\vec{\omega}$ times 2, and another from the complete matrix $\vec{\omega}$:

\begin{align}
\diff{\hat{\I}}{\vec{\omega}} &= \diff{\hat{\I}}{\vec{\Delta}}\left(\diff{\vec{\Delta}}{\text{diag}(\vec{\omega})}\diff{\text{diag}(\vec{\omega})}{\vec{\omega}} + \diff{\vec{\Delta}}{\text{diag}(\vec{\omega})^\top}\diff{\text{diag}(\vec{\omega})^\top}{\vec{\omega}}\right) - 2\times\diff{\hat{\I}}{\vec{\Delta}} \diff{\vec{\Delta}}{\vec{\omega}},\\
&= 2\text{diag}\left(\diff{\hat{\I}}{\vec{\Delta}}\vec{1}_K\right) - 2\diff{\hat{\I}}{\vec{\Delta}}.
\end{align}

We can simplify the factor 2 to get:

\begin{equation}
\diff{\hat{\I}}{\vec{\omega}} = \text{diag}\left(\left[\frac{\vec{\pi}\vec{\pi}^\top}{\sqrt{\vec{\Delta}}}\odot (\vec{1}_{K\times K}-\vec{I}_K)\right]\vec{1}_K\right) - \frac{\vec{\pi}\vec{\pi}^\top}{\sqrt{\vec{\Delta}}}\odot (\vec{1}_{K\times K}-\vec{I}_K).
\end{equation}

This gradient says that on all parts of the matrix except the diagonal, we backpropagate the cross-contribution from $\vec{\omega}$, but sum all this contributions as well on the diagonal. To ease later writings, we introduce $\vec{\Lambda}$:

\begin{equation}
\vec{\Lambda} = \frac{\vec{\pi}\vec{\pi}^\top}{\sqrt{\vec{\Delta}}}\odot (\vec{1}_{K\times K}-\vec{I}_K).
\end{equation}

Thanks to $\vec{\Delta}$ and the cross-product of the vector $\vec{\pi}$, $\vec{\Lambda}$ is positive and symetric.

\subsubsection{Deriving for \texorpdfstring{$\vec{\gamma}$}{gamma}}

Now, we can backpropagate as we did for the OvA MMD, except that $\gamma$ only contributed once to the computation of $\vec{\omega}$. Thus:

\begin{align}
\diff{\hat{\I}}{\vec{\gamma}} &= \vec{\alpha}\diff{\hat{\I}}{\vec{\omega}},\\
&= \vec{\alpha} \left[ \text{diag}\left(\vec{\Lambda}\vec{1}_K\right)- \vec{\Lambda} \right].
\end{align}

\subsubsection{Deriving for \texorpdfstring{$\vec{\alpha}$}{alpha}}

The derivative w.r.t. $\vec{\alpha}$ is a backpropagation through two matrix multiplications: one in $\vec{\gamma}$ and one in $\vec{\omega}$. Hence:

\begin{align}
\diff{\hat{\I}}{\vec{\alpha}} &= \vec{\tilde{\kappa}} \diff{\hat{\I}}{\vec{\gamma}} + \vec{\gamma}\diff{\hat{\I}}{\vec{\omega}},\\
&= \vec{\tilde{\kappa}}\vec{\alpha} \diff{\hat{\I}}{\vec{\omega}} + \vec{\gamma}\diff{\hat{\I}}{\vec{\omega}},\\
&= 2 \vec{\gamma} \diff{\hat{\I}}{\vec{\omega}}.
\end{align}

Thus:

\begin{equation}
\diff{\hat{\I}}{\vec{\alpha}} = 2 \vec{\gamma}\left[ \text{diag}\left(\vec{\Lambda}\vec{1}_K\right)- \vec{\Lambda} \right].
\end{equation}

Note that we can write the first term differently because we multiply the matrix $\vec{\gamma}$ by a diagonal matrix. This is equivalent to doing an element-wise multiplication of $\vec{\gamma}$ by the vector inside the diag function repeated row-wise:

\begin{equation}
\diff{\hat{\I}}{\vec{\alpha}} = 2 \vec{1}_{N\times K}\vec{\Lambda} \odot \vec{\gamma} - 2 \vec{\gamma}\vec{\Lambda}.
\end{equation}

\subsubsection{Deriving for \texorpdfstring{$\vec{\pi}$}{pi}}

The proportions contributed in the final expectation with $\vec{\Delta}$ and in the computations of $\vec{\alpha}$. This looks like what we had in the OvA MMD backpropagation. Therefore:

\begin{align}
\diff{\hat{\I}}{\vec{\pi}} &= \diff{\vec{\pi}^\top\sqrt{\Delta}\vec{\pi}}{\vec{\pi}} - \vec{1}_N^\top \left[\frac{\vec{\alpha}}{\vec{1}_N \vec{\pi}^\top} \odot \diff{\hat{\I}}{\vec{\alpha}}\right],\\
&= 2\vec{\pi}^\top\sqrt{\Delta} - 2\times\vec{1}_{N}^\top \left[\frac{\vec{\alpha}}{\vec{1}_N \vec{\pi}^\top} \odot  \left(\vec{1}_{N\times K}\vec{\Lambda} \odot \vec{\gamma} - \vec{\gamma}\vec{\Lambda}\right)\right].
\end{align}

By noticing that the matrix $\vec{1}_{N\times K}\vec{\Lambda}$ is constant row-wise, we can easily permute the element-wise operation with $\vec{\gamma}$ and perform the matrix multiplication with $\vec{1}_N^\top$ before thanks to factorisation. The element-wise product of $\vec{\alpha}$ with $\vec{\gamma}$ summed over all samples is equal to the diagonal of $\vec{\omega}$. We can rewrite:

\begin{equation}
\diff{\hat{\I}}{\vec{\pi}} = 2\vec{\pi}^\top\sqrt{\Delta} - 2\frac{\text{diag}(\vec{\omega})}{\vec{\pi}^\top} \odot \vec{1}_K\vec{\Lambda} + \frac{2}{\vec{\pi}^\top}\odot\left(\vec{1}_N^\top \left[\vec{\alpha} \odot \vec{\gamma}\vec{\Lambda}\right]\right).
\end{equation}

\subsubsection{Deriving for \texorpdfstring{$\vec{\tau}$}{tau}}

We can finally draw a conclusion to this backpropagation by summing the gradient of the two contributions of $\vec{\tau}$: one from $\vec{\alpha}$ and another one from $\vec{\pi}$:

\begin{align}
\diff{\hat{\I}}{\vec{\tau}}&=\frac{1}{\vec{1}_N\vec{\pi}^\top}\odot \diff{\hat{\I}}{\vec{\alpha}} + \frac{1}{N}\vec{1}_N\diff{\hat{\I}}{\vec{\pi}},\\
\notag&= \frac{2}{\vec{1}_N\vec{\pi}^\top}\odot \left[\vec{\gamma}\odot\vec{1}_{N\times K}\vec{\Lambda} - \vec{\gamma}\vec{\Lambda}\right] + \frac{2}{N}\vec{1}_N\vec{\pi}^\top\sqrt{\vec{\Delta}} \\&\qquad - \frac{2}{N}\vec{1}_N \left[\frac{\text{diag}(\vec{\omega})}{\vec{\pi}^\top} \odot \vec{1}_K\vec{\Lambda}\right] + \frac{2}{N}\vec{1}_N\left[\frac{1}{\vec{\pi}^\top}\odot\left(\vec{1}_N^\top \left[\vec{\alpha} \odot \vec{\gamma}\vec{\Lambda}\right]\right)\right],\\
\notag&= \frac{2}{N}\vec{1}_N\vec{\pi}^\top\sqrt{\vec{\Delta}} +\frac{2}{\vec{1}_N\vec{\pi}^\top} \odot \left[\vec{\gamma}\odot\vec{1}_{N\times K}\vec{\Lambda} - \vec{\gamma}\vec{\Lambda} \right.\\&\qquad\left. -\frac{\vec{1}_N}{N}\left(\text{diag}(\vec{\omega})\odot \vec{1}_K\vec{\Lambda}\right)  + \frac{\vec{1}_{N\times N}}{N}\left(\vec{\alpha}\odot\vec{\gamma}\vec{\Lambda}\right)\right].
\end{align}

\section{Gradients for the Wasserstein GEMINI}
\label{app:wasserstein_computations}

We seek the expression of the gradient of the Wasserstein GEMINI for some output $\vec{\tau}\in\mathbb{R}^{N\times K}$ of some probabilistic model. The matrix $\vec{\tau}$ is therefore row-stochastic. The expression of the GEMINI in the one-vs-all context is then:

\begin{equation}
    \I^\text{ova}_{\mathcal{W}_\delta}(\vec{x};y|\theta) = \E_{y\sim p_\theta(y)}\left[\mathcal{W}_\delta(p_\theta(\vec{x}|y)\|p_\text{data}(x)\right],
\end{equation}
and the one-vs-one variant simply replaces the data distribution with another cluster distribution on which to perform the expectation as well. The distance between the samples is noted $\delta$. During training, the model does not see continuous distribution and only gets batches of samples. Hence, the problem is discretised and the Wasserstein distance can be then evaluated using histogram vectors. We demonstrated~\citep{ohl_generalised_2023} that these histogram vectors consist in a cluster-wise normalisation of the predictions which arises from importance sampling. Thus, the discrete approximation of the Wasserstein GEMINI is:

\begin{equation}
    \hat{\I}^\text{ova}_{\mathcal{W}_\delta}(\vec{x};y|\theta) = \sum_{k=1}^K \pi_k \min_{\vec{P}\in U(\vec{\omega}_k, \vec{1}_N/N)} \sum_{\substack{i=1\\j=1}}^{N,N} \vec{P}_{i,j}\delta(\vec{x}_i,\vec{x}_j),
\end{equation}
where $\vec{P}$ is constrained in a set that forces it to have rows summing to the values of $\vec{\omega}_{\cdot k}$ and columns summing to $\vec{1}_N/N$. The vector $\vec{\omega}_{\cdot k} = \vec{\tau}_{\cdot k} / \sum_{i=1}^N \vec{\tau}_{ik}$ is the normalised cluster predictions.

\subsection{Gradient for the Wasserstein distance}

The new formulation of the discrete Wasserstein distance corresponds to a linear program and is often referred to as the Kantorovich problem. This problem admits the following dual~\citep{peyre_computational_2019}:

\begin{equation}
    \mathcal{W}_\delta(\vec{\omega}_{\cdot 1} \| \vec{\omega}_{\cdot 2}) = \max_{\substack{(\vec{u}, \vec{v})\in \mathbb{R}^N\times\mathbb{R}^N\\\vec{u}_i+\vec{v}_j\leq \delta_{ij},\forall i,j\leq N}} \langle \vec{u},\vec{\omega}_{\cdot 1}\rangle+\langle \vec{v},\vec{\omega}_{\cdot 2}\rangle,
\end{equation}
thanks to the strong duality for linear programs~\cite[p 148, Theorem 4.4]{bertsimas_1997_introduction}. It immediately appears that once we found the optimal "Kantorovich potentials" $\vec{u}^\star$ and $\vec{v}^\star$ for each respective histogram vector $\vec{\omega}_{\cdot 1}$ and $\vec{\omega}_{\cdot 2}$ we can compute the gradient of the distance using these optimal values because we remove the $\max$ term. However, as we want to remain in the simplex, we need to recenter the mass of a gradient and thus subtract the mean of the dual variables:

\begin{equation}
    \diff{\mathcal{W}_\delta(\vec{\omega}_{\cdot 1} \| \vec{\omega}_{\cdot 2})}{\vec{\omega}_{\cdot 1}} = \vec{u}^\star - \sum_{i=1}^N \frac{\vec{u}^\star_i}{N} = \bar{\vec{u}},
\end{equation}
\begin{equation}
    \diff{\mathcal{W}_\delta(\vec{\omega}_{\cdot 1} \| \vec{\omega}_{\cdot 2})}{\vec{\omega}_{\cdot 2}} = \vec{v}^\star - \sum_{i=1}^N \frac{\vec{v}^\star_i}{N} = \bar{\vec{v}}.
\end{equation}

\subsection{Complete gradient for the OvA Wasserstein}

We can now simply unfold the rules of derivation w.r.t. $\vec{\tau}_{ik}$ between the product of the cluster proportions $\pi_k$ and the Wasserstein distance. However, we must take into account that due to the self-normalisation of $\vec{\tau}$ to produce the histogram vectors $\vec{\omega}$, we have to sum its derivative over all normalised samples. Thus:

\begin{align}
    \notag \diff{\hat{\I}}{\vec{\tau}_{ik}}&= \sum_{k^\prime=1}^K \mathcal{W}_\delta(\vec{\omega}_{\cdot k^\prime} \| \vec{1}_N/N) \diff{\pi_{k^\prime}}{\vec{\tau}_{ik}} + \sum_{j=1}^N \pi_{k^\prime}\diff{\mathcal{W}_\delta(\vec{\omega}_{\cdot k^\prime}\|\vec{1}_N/N)}{\vec{\omega}_{jk^\prime}}\diff{\vec{\omega}_{jk^\prime}}{\vec{\tau}_{ik}},\\
    \notag &= \frac{\mathcal{W}_\delta(\vec{\omega}_{\cdot k}\|\vec{1}_N)}{N} + \pi_k \sum_{j=1}^N \bar{\vec{u}}_{jk}\left(\frac{\mathbbm{1}[i==j]}{N\pi_k} - \frac{\vec{\tau_{jk}}}{N^2\pi_k^2}\right),
\end{align}
where $\mathbbm{1}$ is the indicator function resulting from the derivative of the self-normalisation. After summing over all samples, we can conclude that the gradient of the one-vs-all Wasserstein GEMINI w.r.t. model predictions $\vec{\tau}$ is:

\begin{equation}
    \diff{\hat{\I}^\text{ova}_{\mathcal{W}_\delta}}{\vec{\tau}_{ik}} = \frac{\mathcal{W}_\delta(\vec{\omega}_{\cdot k}\|\vec{1}_N)}{N} + \frac{\bar{\vec{u}}_{ik}}{N} - \frac{\langle \bar{\vec{u}}_{\cdot k}, \vec{\tau}_k \rangle}{N^2 \pi_k}.
\end{equation}

\subsection{Complete gradient for the OvO Wasserstein}

The demonstration follows the same rules as before. We add as well the fact that the Wasserstein distance is symmetric, and hence its gradient is as well so we can permute the names $\bar{\vec{u}}$ and $\bar{\vec{v}}$ when changing $\mathcal{W}_\delta(\vec{\omega}_1\|\vec{\omega}_2)$ for $\mathcal{W}_\delta(\vec{\omega}_2\|\vec{\omega}_1)$. Therefore, we sum twice the gradients of the Wasserstein distances, as well as twice the gradients for the proportions due to the symmetric nature of this function. We can thus arrive to a final gradient that is very similar to the OvO scenario with an additional summing over adversarial proportions:

\begin{equation}
    \diff{\hat{\I}^\text{ovo}_{\mathcal{W}_\delta}}{\vec{\tau}_{ik}} = \sum_{k^\prime=1}^K 2\frac{\pi_k\mathcal{W}_\delta(\vec{\omega}_{\cdot k} \| \vec{\omega}_{\cdot k^\prime})}{N} + 2\frac{\pi_{k^\prime}\bar{\vec{u}}_{j,k/k^\prime}}{N} - 2\frac{\langle \bar{\vec{u}}_{\cdot k/k^\prime} \vec{\tau}_{k}\rangle}{N^2 \pi_k}.
\end{equation}

Notice that we detailed in subscript for which Wasserstein evaluation a dual variable emerges using the notation $k/k^\prime$. Since the one-vs-one GEMINI makes $K^2$ distance evaluation, we have $K^2$ dual variables as well when removing duplicate dual variables due to symmetry.

\section{Examples of code snippets with the package GemClus}
\label{app:gemclus_details}

GemClus is implemented to respect as much as possible the scikit-learn~\cite{pedregosa_scikit-learn_2011} naming conventions. For instance, Listing~\ref{list:gemclus_example} shows how a the logistic regression can be trained with an MMD GEMINI on the breast cancer dataset. The Listing~\ref{list:code_experiment} shows how the numerical experiments from Section~\ref{ssec:numerical_experiments} can be run.

We also want to extend this package to other discriminative clustering methods for potentially small-scale datasets. Therefore, we include an implementation of the regularised mutual information (RIM) model by \citet{krause_discriminative_2010} as shown in Listing~\ref{list:rim_example} because we consider this model to be one of the very first proposed in the domain yet find few satisfying implementations. In this sense, we included as well (and as can be noted in Listing~\ref{list:code_experiment} functions for generating relevant synthetic datasets for clustering.

\begin{lstlisting}[float, language=PythonPlus, style=colorEX, caption=\centering An example of \emph{gemclus} loading and data clustering
\label{list:gemclus_example}]
from gemclus.sparse import SparseLinearMMD
from sklearn.datasets import load_breast_cancer
# load data
X, _ = load_breast_cancer(return_X_y=True)
# Create simple logistic regression model and do clustering
y_pred = SparseLinearMMD(n_clusters=2).fit_predict(X)
\end{lstlisting}

\begin{lstlisting}[float,language=PythonPlus, style=colorEX, caption=\centering An example of sparse GEMINI model fitting the 5th scenario of the synthetic datasets\label{list:code_experiment}]
from gemclus.sparse import SparseMLPMMD
from gemclus.data import celeux_one

# Generate the data according to the 5th scenario
X,y = celeux_one(n=300, p=95, mu=1.7)
# Prepare the model: MLP with the OvA MMD-GEMINI for 3 clusters
model = SparseMLPMMD(n_clusters=3)
# Progressively increase the penalty until all features are removed
# res contains the history of feature selection and best model weights
res = model.path(X)
\end{lstlisting}

\begin{lstlisting}[float, language=PythonPlus, style=colorEX,caption=\centering The package \emph{gemclus} incorporates as well the basic logistic regression with regularised mutual information by~\citet{krause_discriminative_2010}\label{list:rim_example}]
from gemclus.linear import RIM
y_pred = RIM(n_clusters=3).fit_predict(X)
\end{lstlisting}

\section{Results on the synthetic datasets for the one-vs-all GEMINIs}
\label{app:synthetic_wasserstein}

\begin{table*}
\centering
\caption{Performances of Sparse GEMINI using the one-vs-all objectives on the synthetic datasets.}
\label{tab:synthetic_ova}
    \vskip 0.1in
\subfloat[][ARI scores (greater is better)] {
	\begin{tabular}{c c c c c}
	\toprule
	&\multicolumn{2}{c}{MMD-GEMINI}&\multicolumn{2}{c}{Wasserstein-GEMINI}\\
	\cmidrule{2-3}\cmidrule{4-5}
	&Logistic&MLP&Logistic&MLP\\
	\midrule
	S1&0.08\std{0.08}&\textbf{0.11\std{0.12}}&0.03\std{0.05}&0.08\std{0.09}\\
	S2&\textbf{0.46\std{0.11}}&\textbf{0.47\std{0.11}}&\textbf{0.46\std{0.13}}&0.41\std{0.15}\\
	S3*&\textbf{0.23\std{0.07}}&\textbf{0.22\std{0.05}}&0.13\std{0.07}&0.08\std{0.06}\\
	S4&0.43\std{0.05}&0.45\std{0.05}&0.56\std{0.17}&\textbf{0.74\std{0.22}}\\
	S5&0.44\std{0.05}&\textbf{0.58\std{0.11}}&0.52\std{0.16}&0.47\std{0.18}\\
	\cmidrule{2-5}
	D2&0.53\std{0.06}&0.55\std{0.03}&\textbf{0.57\std{0.02}}&0.54\std{0.03}\\
	\bottomrule
	\end{tabular}
}\hfill
\subfloat[][VSER scores (lower is better)]{
	\begin{tabular}{c c c c c}
	\toprule
	&\multicolumn{2}{c}{MMD-GEMINI}&\multicolumn{2}{c}{Wasserstein-GEMINI}\\
	\cmidrule{2-3}\cmidrule{4-5}
	&Logistic&MLP&Logistic&MLP\\
	\midrule
	S1&\textbf{0.38\std{0.14}}&\textbf{0.37\std{0.11}}&0.58\std{0.13}&0.55\std{0.10}\\
	S2*&\textbf{0.04\std{0.04}}&0.10\std{0.06}&0.16\std{0.08}&0.25\std{0.11}\\
	S3*&\textbf{0.06\std{0.08}}&0.19\std{0.11}&0.20\std{0.13}&0.68\std{0.16}\\
	S4&\textbf{0.00\std{0.00}}&\textbf{0.00\std{0.00}}&0.03\std{0.03}&\textbf{0.00\std{0.00}}\\
	S5&\textbf{0.00\std{0.00}}&\textbf{0.00\std{0.00}}&0.02\std{0.02}&0.08\std{0.07}\\
	\cmidrule{2-5}
	D2&0.29\std{0.00}&0.30\std{0.04}&0.29\std{0.00}&0.29\std{0.03}\\
	\bottomrule
	\end{tabular}
}\hfill
\subfloat[][CVR scores (greater is better)]{
	\begin{tabular}{c c c c c}
	\toprule
	&\multicolumn{2}{c}{MMD-GEMINI}&\multicolumn{2}{c}{Wasserstein-GEMINI}\\
	\cmidrule{2-3}\cmidrule{4-5}
	&Logistic&MLP&Logistic&MLP\\
	\midrule
	S1&0.56\std{0.20}&0.63\std{0.31}&0.65\std{0.21}&\textbf{0.71\std{0.17}}\\
	S2&\textbf{0.93\std{0.10}}&0.84\std{0.12}&\textbf{0.91\std{0.10}}&0.81\std{0.12}\\
	S3&0.94\std{0.16}&\textbf{0.98\std{0.06}}&0.66\std{0.23}&\textbf{0.96\std{0.08}}\\
	S4&\textbf{1.00\std{0.00}}&\textbf{1.00\std{0.00}}&0.95\std{0.09}&0.99\std{0.04}\\
	S5&\textbf{1.00\std{0.00}}&\textbf{1.00\std{0.00}}&0.94\std{0.09}&0.95\std{0.09}\\
	\cmidrule{2-5}
	D2&0.00\std{0.00}&0.00\std{0.00}&0.00\std{0.00}&0.00\std{0.00}\\
	\bottomrule
	\end{tabular}
}
\end{table*}

We provide here a complement on the numerical experiments from section~\ref{ssec:numerical_experiments} with the performances of Sparse GEMINI when using the one-vs-all GEMINIs in Table~\ref{tab:synthetic_ova}.

\section{Distribution of numerical experiments scores}
\label{app:boxplots}

\begin{figure}
    \centering
    \subfloat[ARI]{
        \includegraphics[width=0.8\linewidth]{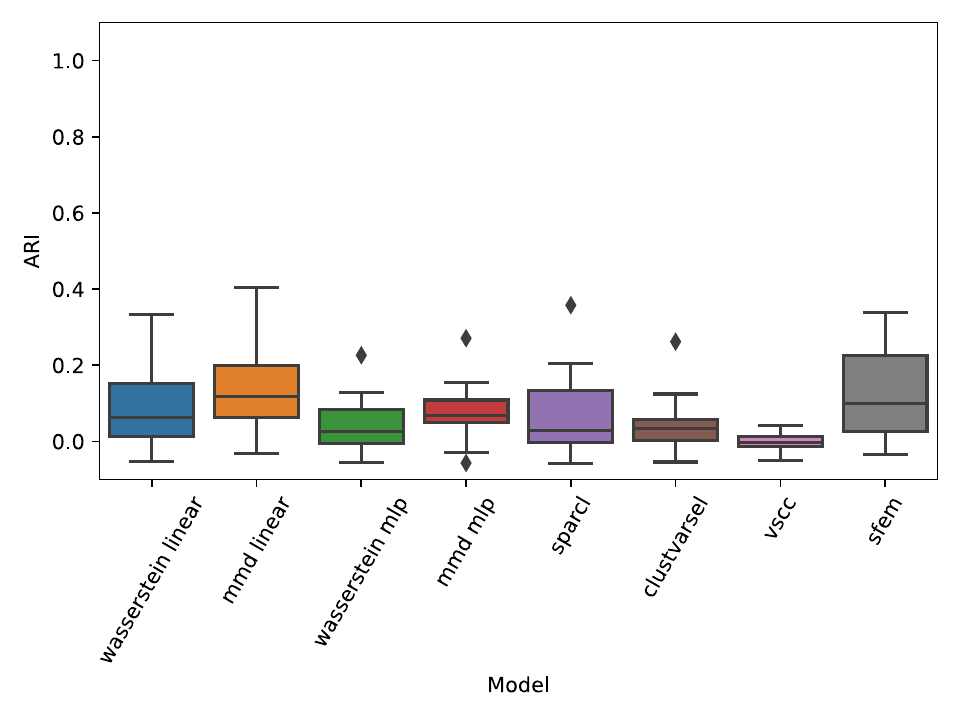}
    }\\
    \subfloat[VSER]{
        \includegraphics[width=0.8\linewidth]{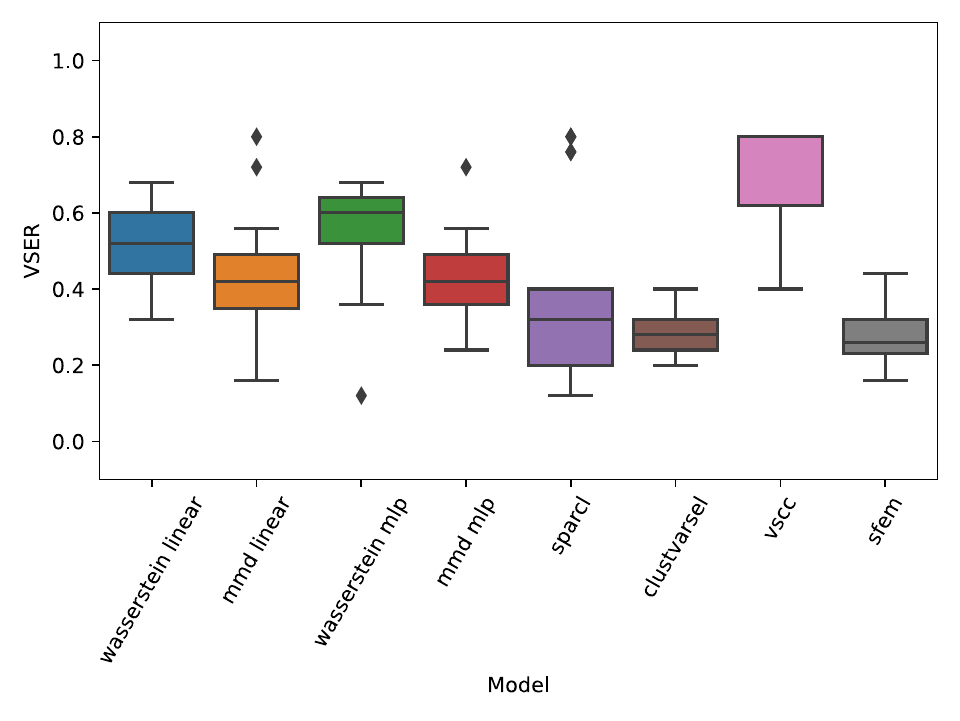}    
    }
    \caption{Box plots of one-vs-one methods against baseline on the dataset S1.}
    \label{fig:1s1_boxplot}
\end{figure}

\begin{figure}
    \centering
    \subfloat[ARI]{
        \includegraphics[width=0.8\linewidth]{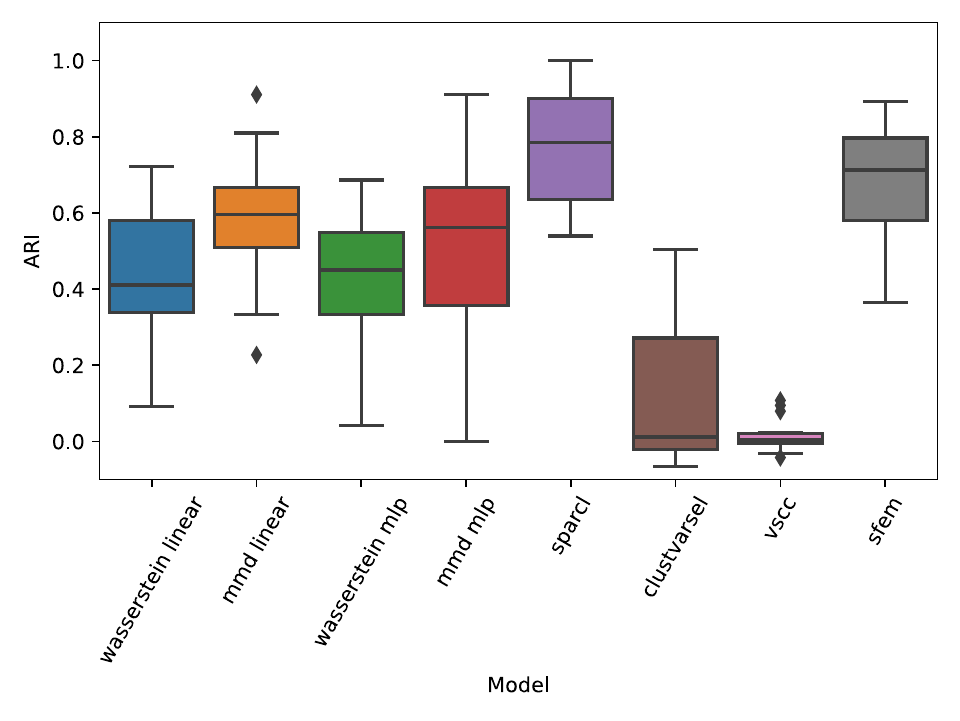}
    }\\
    \subfloat[VSER]{
        \includegraphics[width=0.8\linewidth]{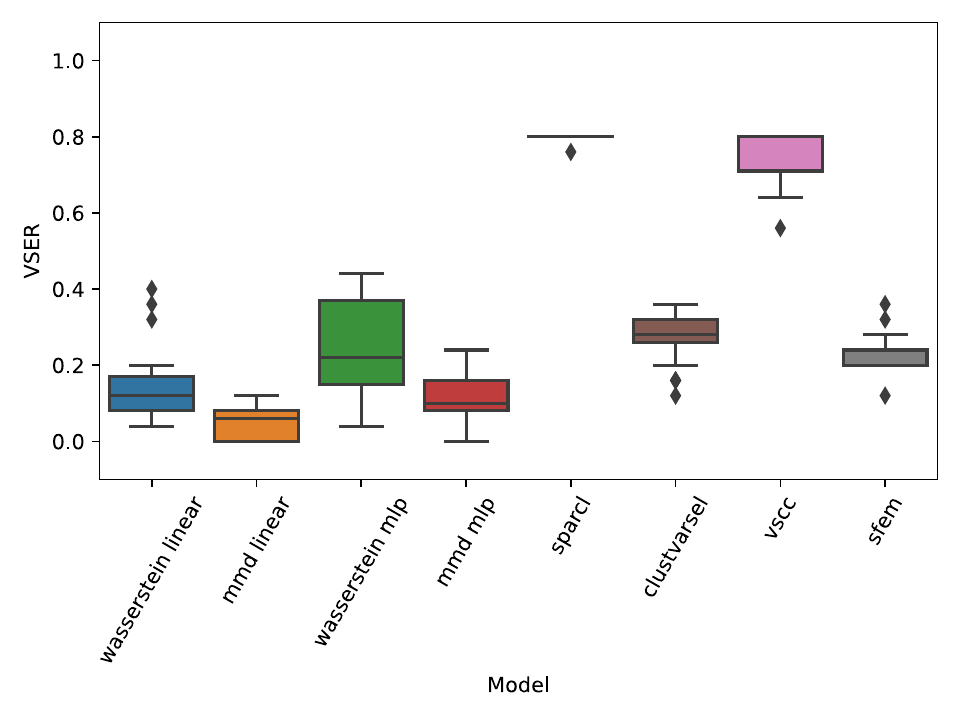}    
    }
    \caption{Box plots of one-vs-one methods against baseline on the dataset S2.}
    \label{fig:1s2_boxplot}
\end{figure}

\begin{figure}
    \centering
    \subfloat[ARI]{
        \includegraphics[width=0.8\linewidth]{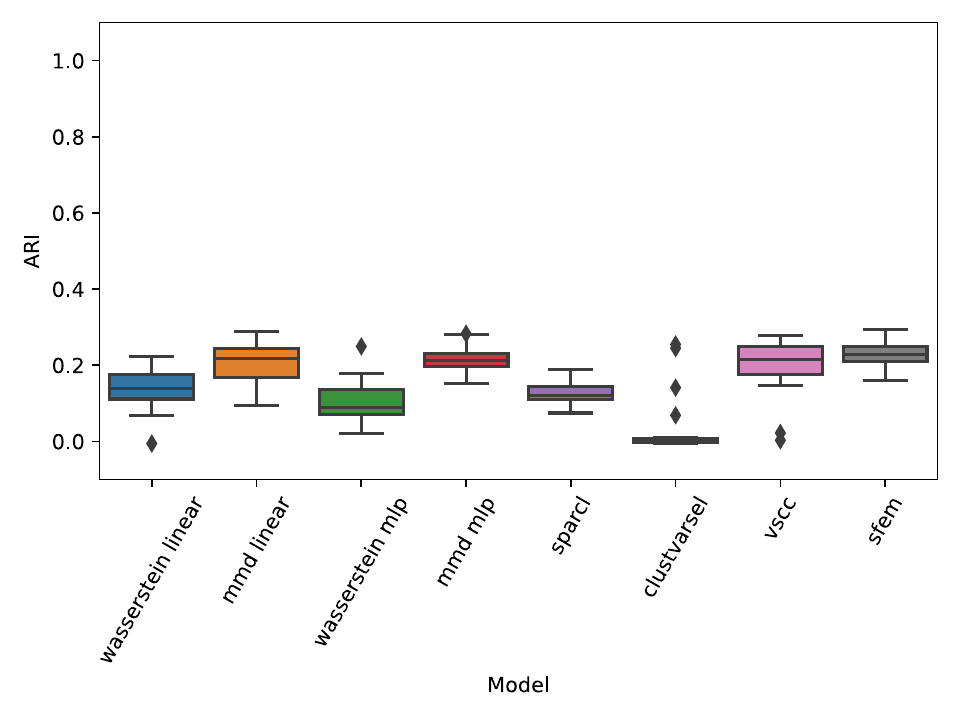}
    }\\
    \subfloat[VSER]{
        \includegraphics[width=0.8\linewidth]{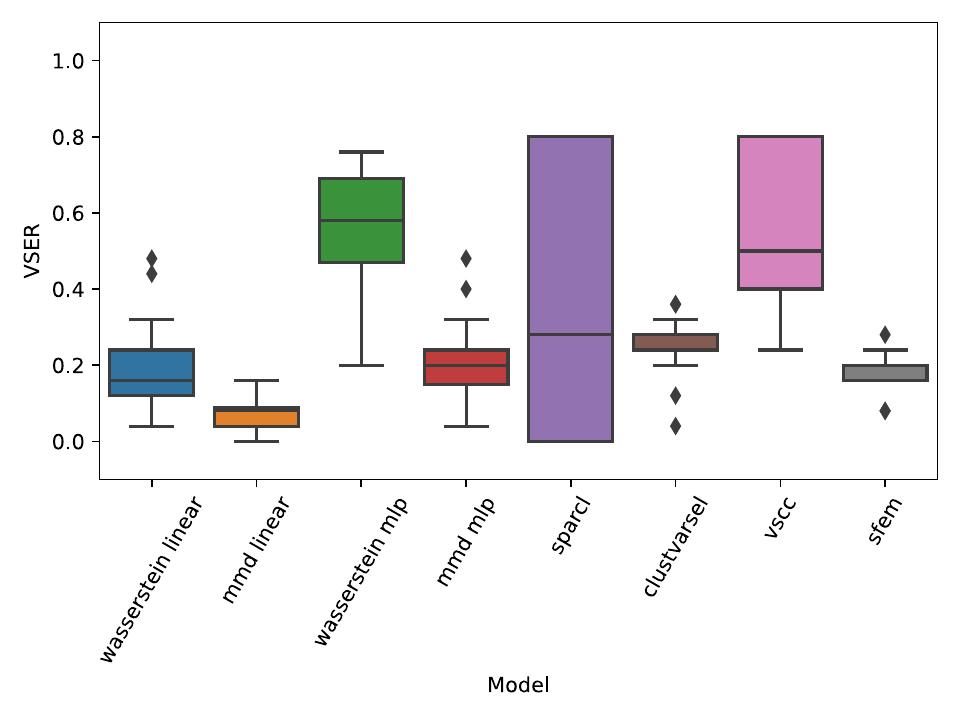}    
    }
    \caption{Box plots of one-vs-one methods against baseline on the dataset S3.}
    \label{fig:1s3_boxplot}
\end{figure}

\begin{figure}
    \centering
    \subfloat[ARI]{
        \includegraphics[width=0.8\linewidth]{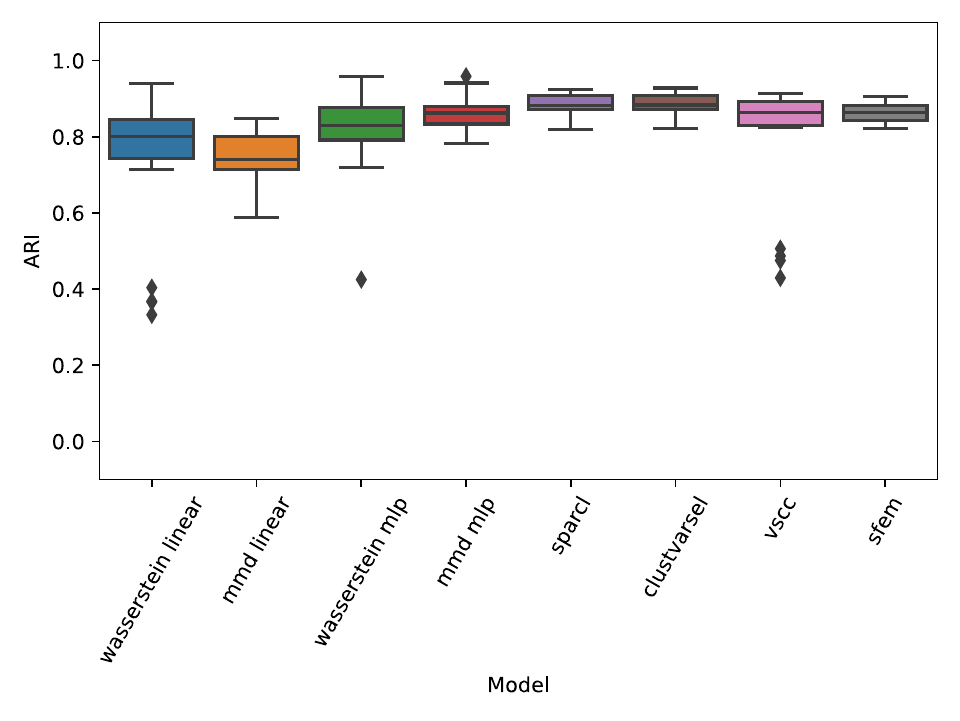}
    }\\
    \subfloat[VSER]{
        \includegraphics[width=0.8\linewidth]{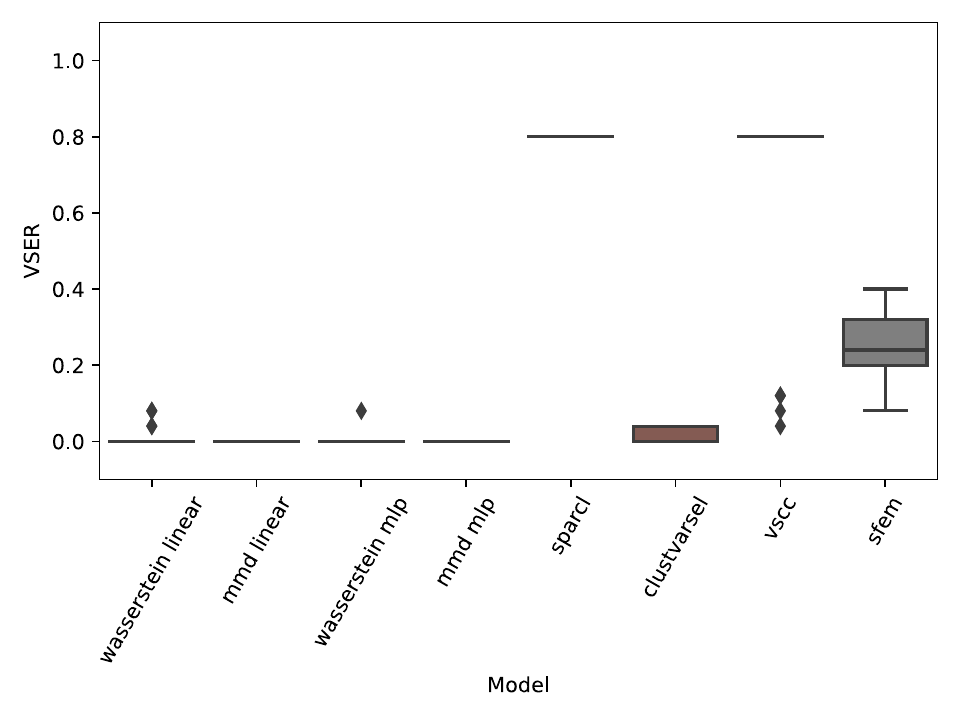}    
    }
    \caption{Box plots of one-vs-one methods against baseline on the dataset S4.}
    \label{fig:1s4_boxplot}
\end{figure}

\begin{figure}
    \centering
    \subfloat[ARI]{
        \includegraphics[width=0.8\linewidth]{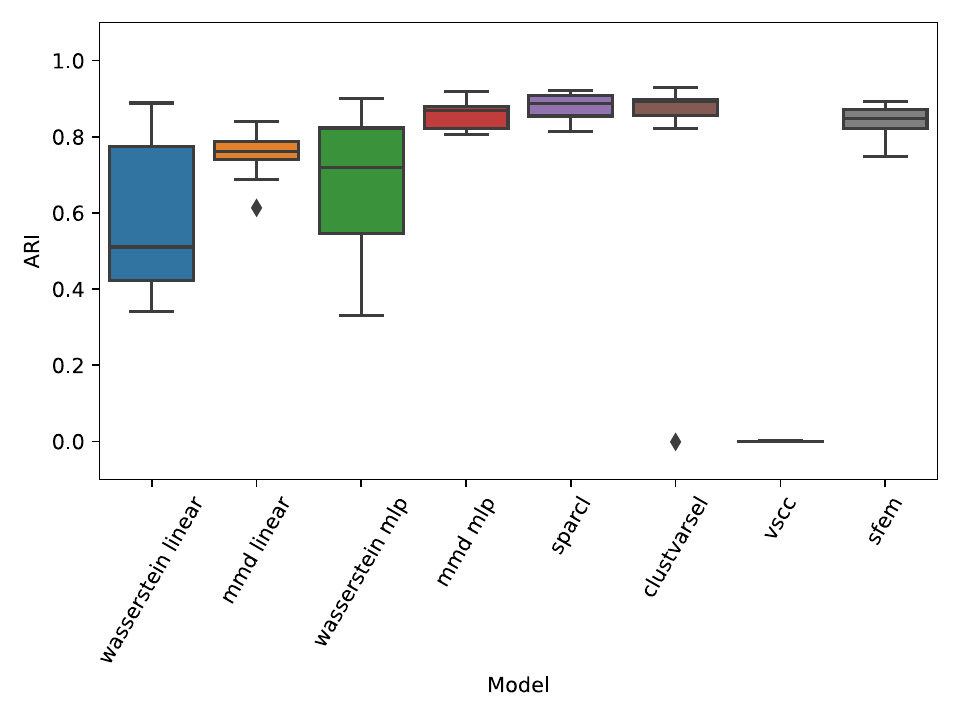}
    }\\
    \subfloat[VSER]{
        \includegraphics[width=0.8\linewidth]{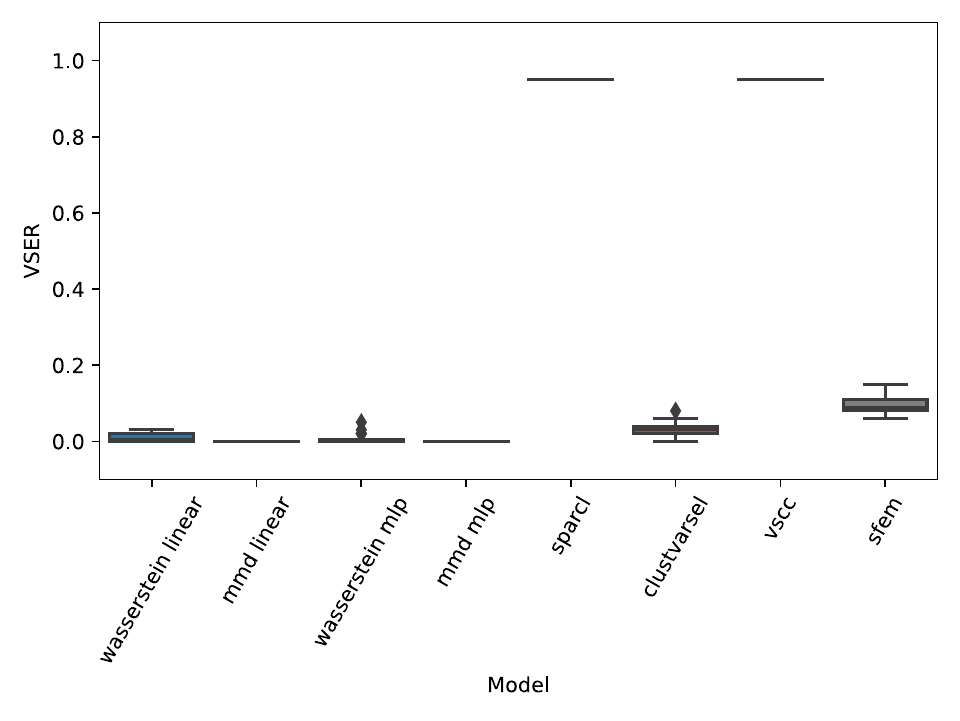}    
    }
    \caption{Box plots of one-vs-one methods against baseline on the dataset S5.}
    \label{fig:1s5_boxplot}
\end{figure}

We provide here additional details regarding the distribution of the ARI and VSER scores from Table~\ref{tab:synthetic_datasets_results} for the one-vs-one models in static training. These scores are displayed with box plots respectively in figures~\ref{fig:1s1_boxplot}, \ref{fig:1s2_boxplot}, \ref{fig:1s3_boxplot}, \ref{fig:1s4_boxplot} and \ref{fig:1s5_boxplot} for the scenarios 1, 2, 3, 4, 5 of the first synthetic dataset.


\end{appendices}

\clearpage

%%%%%%%%%%%%%%%%%%%%%%%%%%%%%%%%%%%%%%%%%%%%%%%%%%%%%%%%%%%%%%%%%%%%%%%%%
% BIBLIOGRAPHY
%%%%%%%%%%%%%%%%%%%%%%%%%%%%%%%%%%%%%%%%%%%%%%%%%%%%%%%%%%%%%%%%%%%%%%%%%
\bibliography{bib}

\end{document}